\documentclass[runningheads]{llncs}
\usepackage{graphicx}

\usepackage{tikz}
\usepackage{comment}
\usepackage{amsmath,amssymb} 
\usepackage{color}

\usepackage[accsupp]{axessibility}  

\usepackage{tikz}
\usepackage{comment}
\usepackage{amsmath,amssymb} 
\usepackage{color}
\usepackage{epsfig}
\usepackage{graphicx}
\usepackage{amssymb}
\usepackage{caption}
\usepackage{subcaption}
\usepackage{multirow}
\usepackage{mathrsfs}
\usepackage{calligra}
\usepackage{dsfont}
\usepackage{pifont}
\usepackage{pbox}
\usepackage{url}
\usepackage[misc]{ifsym}
\usepackage{tabularx,booktabs}
\usepackage[export]{adjustbox}

\usepackage[accsupp]{axessibility}  

\newlength\savewidth\newcommand\shline{\noalign{\global\savewidth\arrayrulewidth
  \global\arrayrulewidth 1pt}\hline\noalign{\global\arrayrulewidth\savewidth}}

\newcommand{\RANKSEG}{{\textcolor{blue}{+ RankSeg}}}

\makeatletter
\def\@fnsymbol#1{\ensuremath{\ifcase#1\or \dagger\or \ddagger\or
\mathsection\or \mathparagraph\or \|\or **\or \dagger\dagger
\or \ddagger\ddagger \else\@ctrerr\fi}}
\makeatother

\begin{document}
\pagestyle{headings}
\mainmatter
\def\ECCVSubNumber{100}  

\title{\textsc{RankSeg}: Adaptive Pixel Classification with Image Category Ranking for Segmentation}

\titlerunning{Adaptive Pixel Classification with Category Ranking for Segmentation}
%
\author{Haodi He\inst{1}\thanks{Equal contribution.} \and
Yuhui Yuan\inst{3}$^\dagger$ \and
Xiangyu Yue\inst{2}\and
Han Hu\inst{3}}
\authorrunning{Haodi He, Yuhui Yuan, Xiangyu Yue, Han Hu.}
%
\institute{University of Science and Technology of China\and UC Berkeley\and
Microsoft Research Asia\\
\email{\Letter\;\{yuhui.yuan,hanhu\}@microsoft.com}}

\maketitle

\begin{abstract}
The segmentation task has traditionally been formulated as a complete-label\footnote{We use the term ``complete label'' to represent the set of all predefined categories in the dataset.} pixel classification task to predict a class for each pixel from a fixed number of predefined semantic categories shared by all images or videos.
Yet, following this formulation, standard architectures will inevitably encounter various challenges under more realistic settings where the scope of categories scales up (e.g., beyond the level of $1\rm{k}$).
On the other hand, in a typical image or video, only a few categories, i.e., a small subset of the complete label are present.
Motivated by this intuition, in this paper, we propose to decompose segmentation into two sub-problems: (i) image-level or video-level multi-label classification and (ii) pixel-level rank-adaptive selected-label classification. Given an input image or video, our framework first conducts multi-label classification over the complete label, then sorts the complete label and selects a small subset according to their class confidence scores.
We then use a rank-adaptive pixel classifier to perform the pixel-wise classification over only the selected labels, which uses a set of rank-oriented learnable temperature parameters to adjust the pixel classifications scores.
Our approach is conceptually general and can be used to improve various existing
segmentation frameworks by simply using a lightweight
multi-label classification head and rank-adaptive pixel classifier.
We demonstrate the effectiveness of our framework with competitive experimental results across four tasks, including image semantic segmentation, image panoptic segmentation, video instance segmentation, and video semantic segmentation.
Especially, with our RankSeg, Mask$2$Former gains +$0.8\%$/+$0.7\%$/+$0.7\%$ on ADE$20$K panoptic segmentation/YouTubeVIS $2019$ video instance segmentation/VSPW video semantic segmentation benchmarks respectively.
Code is available at:
{\footnotesize\url{{https://github.com/openseg-group/RankSeg}}}
\end{abstract}

\section{Introduction}

\begin{figure}[t]
\begin{subfigure}[b]{0.22\linewidth}
\centering
\includegraphics[width=\linewidth,height=17.5mm]{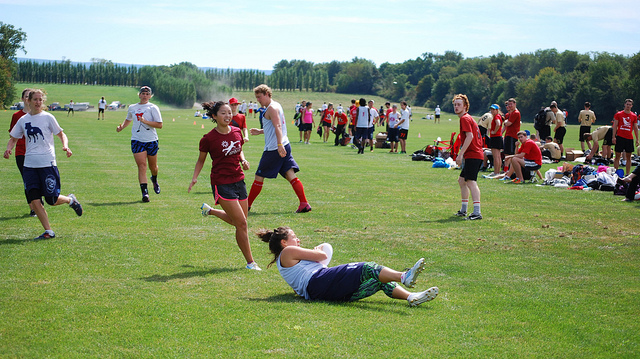}
\caption{\small{Image}}
\end{subfigure}%
\begin{subfigure}[b]{0.22\linewidth}
\centering
\includegraphics[width=\linewidth,height=17.5mm]{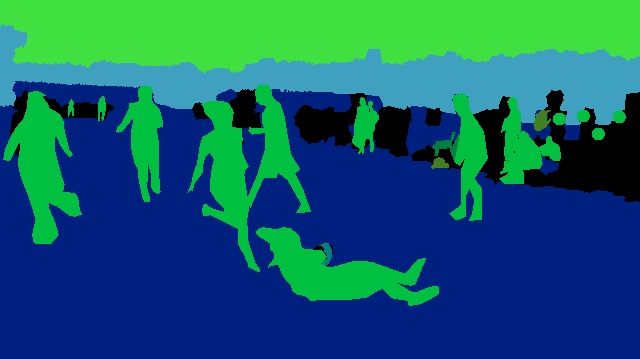}
\caption{\small{Ground-truth}}
\end{subfigure}
\begin{subfigure}[b]{0.54\textwidth}
\includegraphics[width=1\textwidth,height=17.5mm]{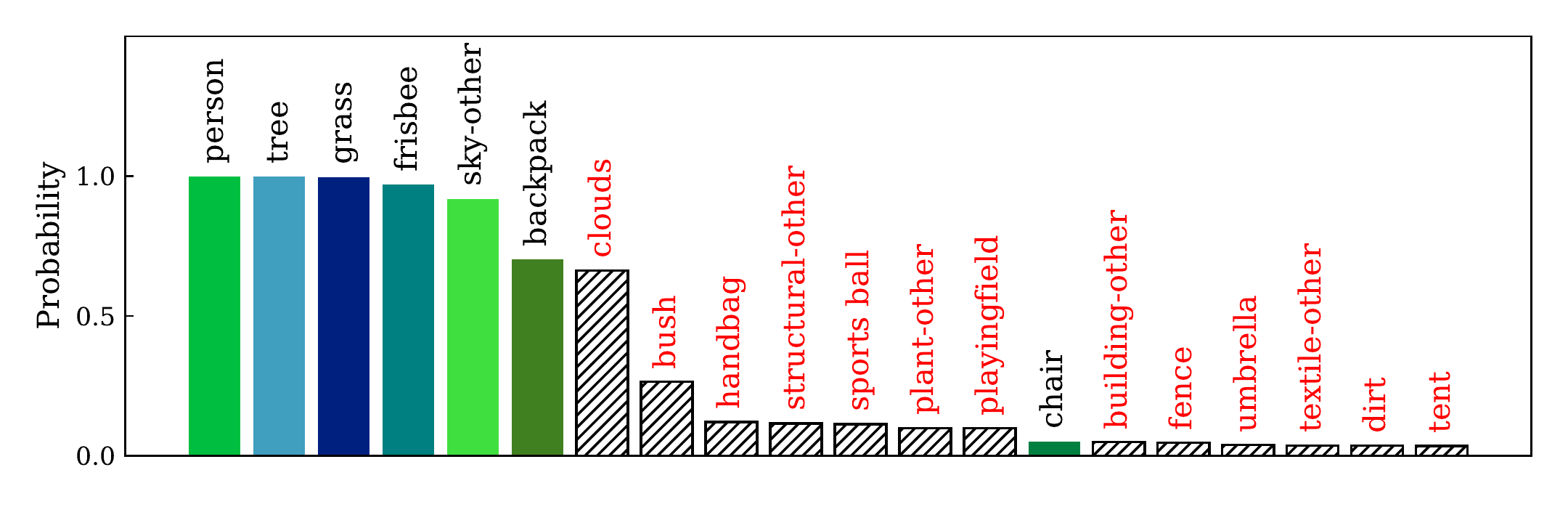}
\caption{\small{Multi-label image classification prediction}}
\end{subfigure}
\caption{\small{\textbf{Illustrating the motivation of exploiting the multi-label image classification}: (a) An example image selected from COCO-Stuff. (b) The ground-truth segmentation map consisting of $7$ classes. (c) The histogram of existence probability over the selected top $20$ categories sorted by their confidence scores, which are predicted with our method. Our method only needs to identify the label of each pixel from these selected $20$ categories instead of all $171$ categories. The names of true/false positive categories are marked with \color{black}{black}/{\color{
red}{red}} color, respectively. The bars associated with true positive categories share the same color as the ones adopted in the ground-truth segmentation map.}}
\label{fig:ml_exp_case}
\end{figure}

Image and video segmentation, i.e., partitioning images or video frames into multiple meaningful segments, is a fundamental computer vision research topic
that has wide applications including autonomous driving, surveillance system,
and augmented reality.
Most recent efforts have followed the path of fully convolutional networks~\cite{long2015fully} and proposed various advanced improvements, e.g.,
high-resolution representation learning~\cite{pohlen2017full,WangSCJDZLMTWLX19},
contextual representation aggregation~\cite{chen2017rethinking,fu2019dual,huang2019ccnet,zhao2017pyramid,yuan2020object},
boundary refinement~\cite{takikawa2019gated,kirillov2020pointrend,yuan2020segfix},
and vision transformer architecture designs~\cite{liu2021swin,ranftl2021vision,yuan2021hrformer,zheng2021rethinking}.

Most of the existing studies formulate the image and video segmentation problem as a complete-label pixel classification task.
In the following discussion, we'll take image semantic segmentation as an example for convenience.
For example, image semantic segmentation needs to select the label of each pixel from the complete label\footnote{We use ``label'', ``category'', and ``class'' interchangeably.} set that is predefined in advance.
However, it is unnecessary to consider the complete label set
for every pixel in each image as most standard images only consist of objects belonging to a few categories.
Figure~\ref{fig:num_label_histogram} plots the statistics on the percentage of images that contain
no more than the given class number in the entire dataset vs.
the number of classes that appear within each image.
Accordingly, we can see that $100.00\%$, $99.99\%$, $99.14\%$, and $99.85\%$ of images contain less than $25$ categories on PASCAL-Context~\cite{mottaghi2014role}, COCO-Stuff~\cite{caesar2018coco}, ADE$20$K-Full~\cite{cheng2021per,zhou2017scene}, and COCO+LVIS~\cite{gupta2019lvis,jain2021scaling} while each of them contains $60$, $171$, $847$, and $1,284$ predefined semantic categories respectively.
Besides, Figure~\ref{fig:ml_exp_case} shows an example image that only contains $7$ classes while the complete label set consists of $171$ predefined categories.

To take advantage of the above observations, we propose to
re-formulate the segmentation task into two sub-problems
including multi-label image/video classification and
rank-adaptive pixel classification over a subset of selected labels.
To verify the potential benefits of our method,
we investigate the gains via exploiting the ground-truth multi-label
of each image or video. In other words, the pixel classifier only needs to
select the category of each pixel from a collection of categories presented
in the current image or video, therefore, we can filter out all other categories that do not appear.
Figure~\ref{fig:gt_rankseg} summarizes the comparison results based on Segmenter~\cite{strudel2021} and Mask$2$Former~\cite{cheng2021masked}.
We can see that the segmentation performance is significantly improved given the ground-truth multi-label prediction.
In summary, multi-label classification
is an important but long-neglected sub-problem on the path toward more accurate segmentation.

\definecolor{electricviolet}{rgb}{0.4, 0.23, 0.7}
\begin{figure*}[t]
\begin{minipage}[t]{1\linewidth}
\centering
\begin{subfigure}[b]{0.235\textwidth}
\includegraphics[width=\textwidth]{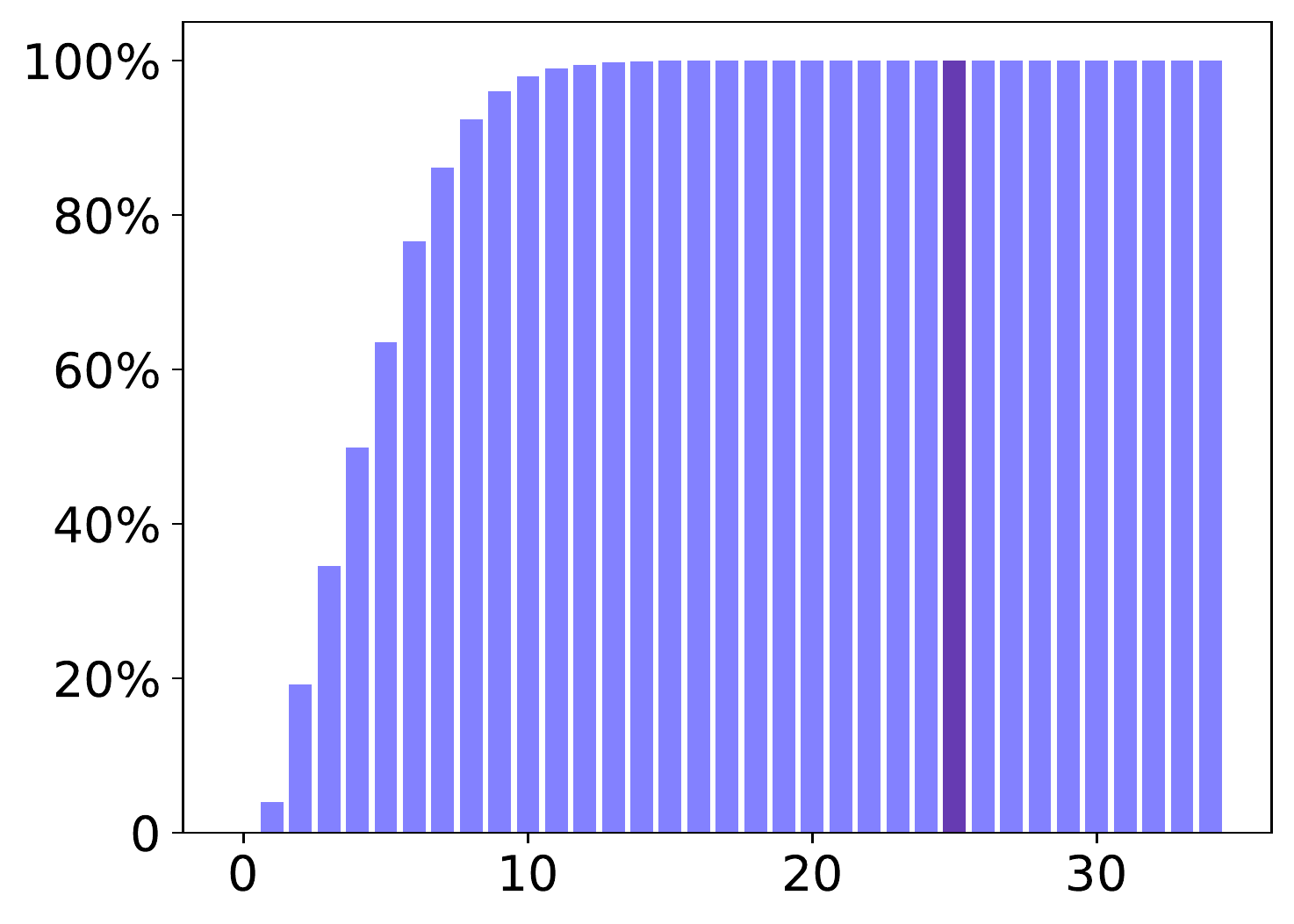}
\caption{\tiny{PASCAL-Context~\cite{mottaghi2014role}}}
\end{subfigure}
\begin{subfigure}[b]{0.235\textwidth}
{\includegraphics[width=\textwidth]{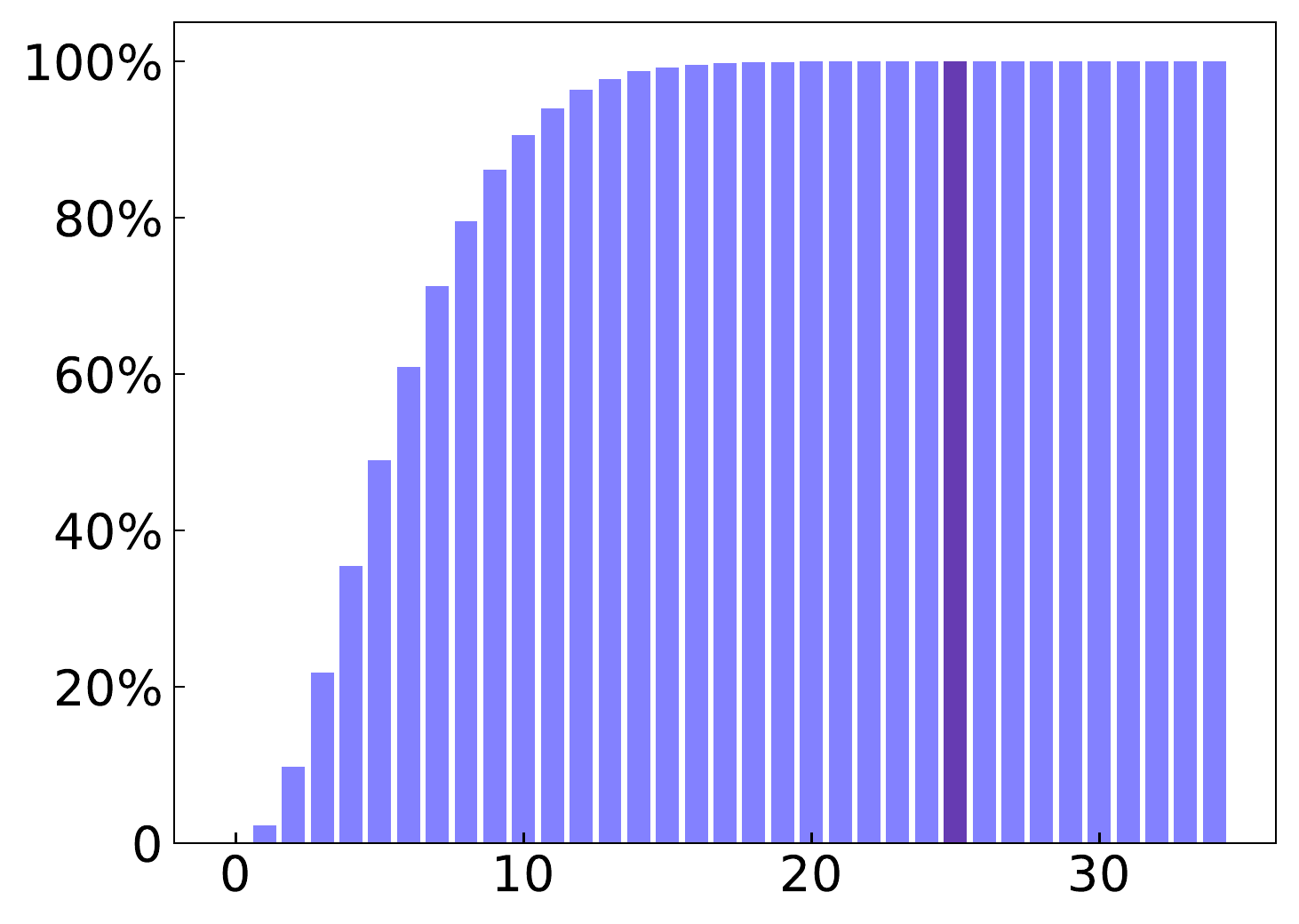}}
\caption{\tiny{COCO-Stuff}~\cite{caesar2018coco}}
\end{subfigure}
\begin{subfigure}[b]{0.235\textwidth}
{\includegraphics[width=\textwidth]{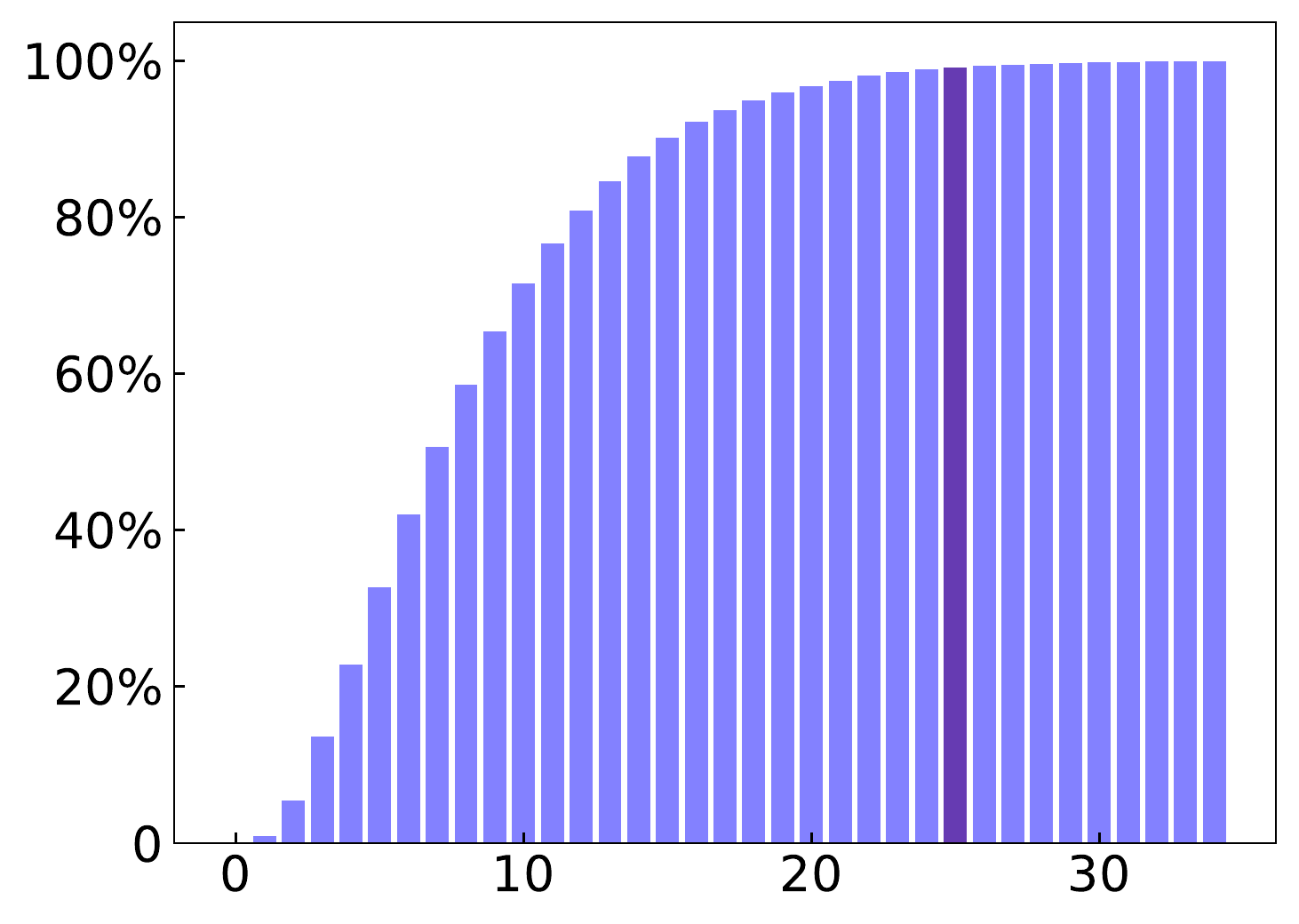}}
\caption{\tiny{ADE$20$K-Full}~\cite{zhou2017scene,cheng2021per}}
\end{subfigure}
\begin{subfigure}[b]{0.235\textwidth}
{\includegraphics[width=\textwidth]{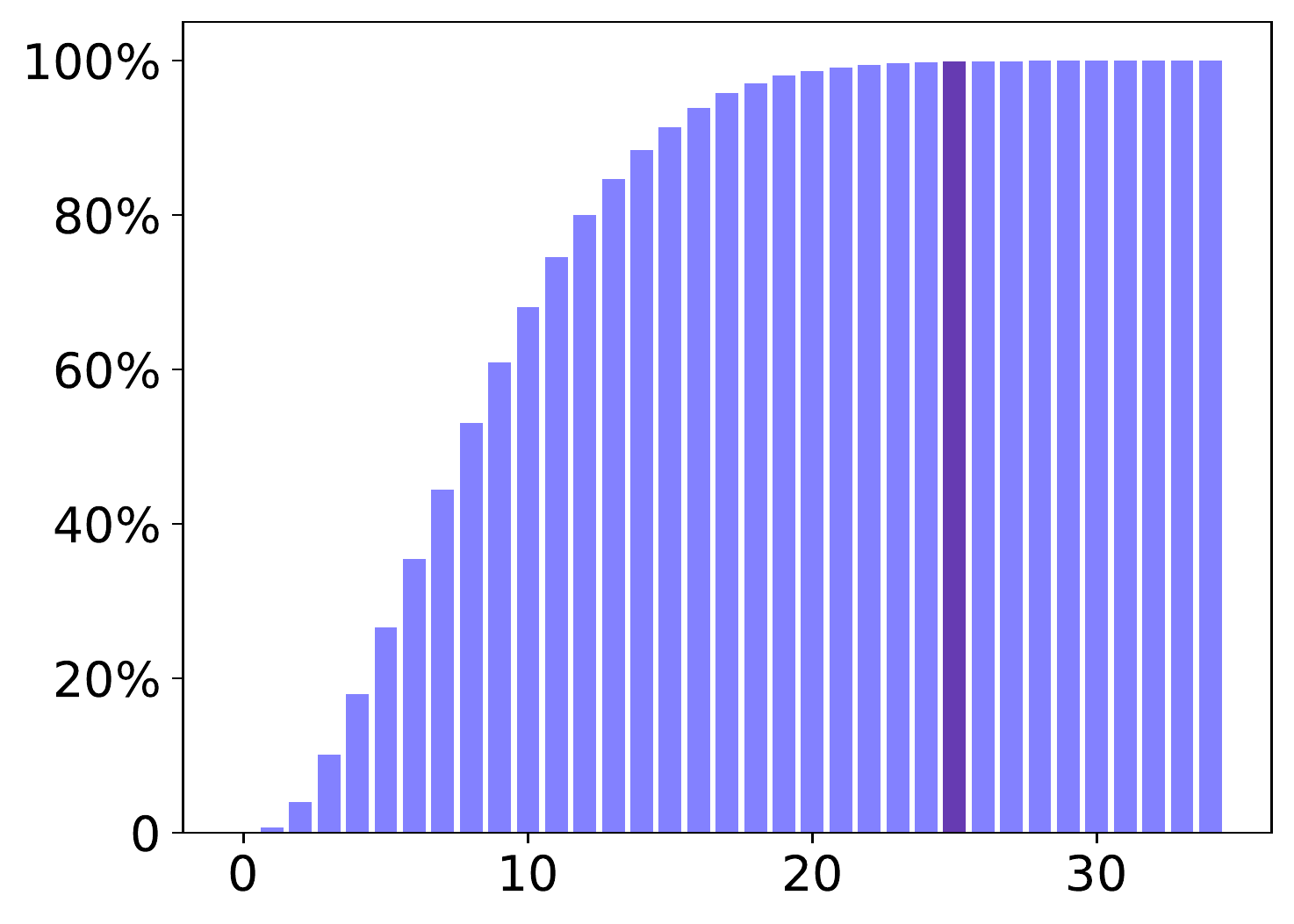}}
\caption{\tiny{COCO+LVIS}~\cite{gupta2019lvis,jain2021scaling}}
\end{subfigure}
\caption{\small{\textbf{Illustrating the cumulative distribution
of the number of images that contain no more than the given class number}:
The $x$-axis represents the class number presented in the image
and the $y$-axis represents the percentage of images that contain
no more than the given class number in the entire dataset.
We plot the cumulative distribution on four benchmarks.
We can see that more than $99\%$ of all images only contain less than $25$ categories on four benchmarks, which are marked with {\color{electricviolet}{dark purple}} bars.
The above four benchmarks contain $60$, $171$, $847$, and $1,284$ predefined semantic categories respectively.}}
\label{fig:num_label_histogram}
\end{minipage}
\begin{minipage}[t]{1\linewidth}
\centering
\begin{subfigure}[b]{0.465\textwidth}
{\includegraphics[width=\textwidth,height=25mm]{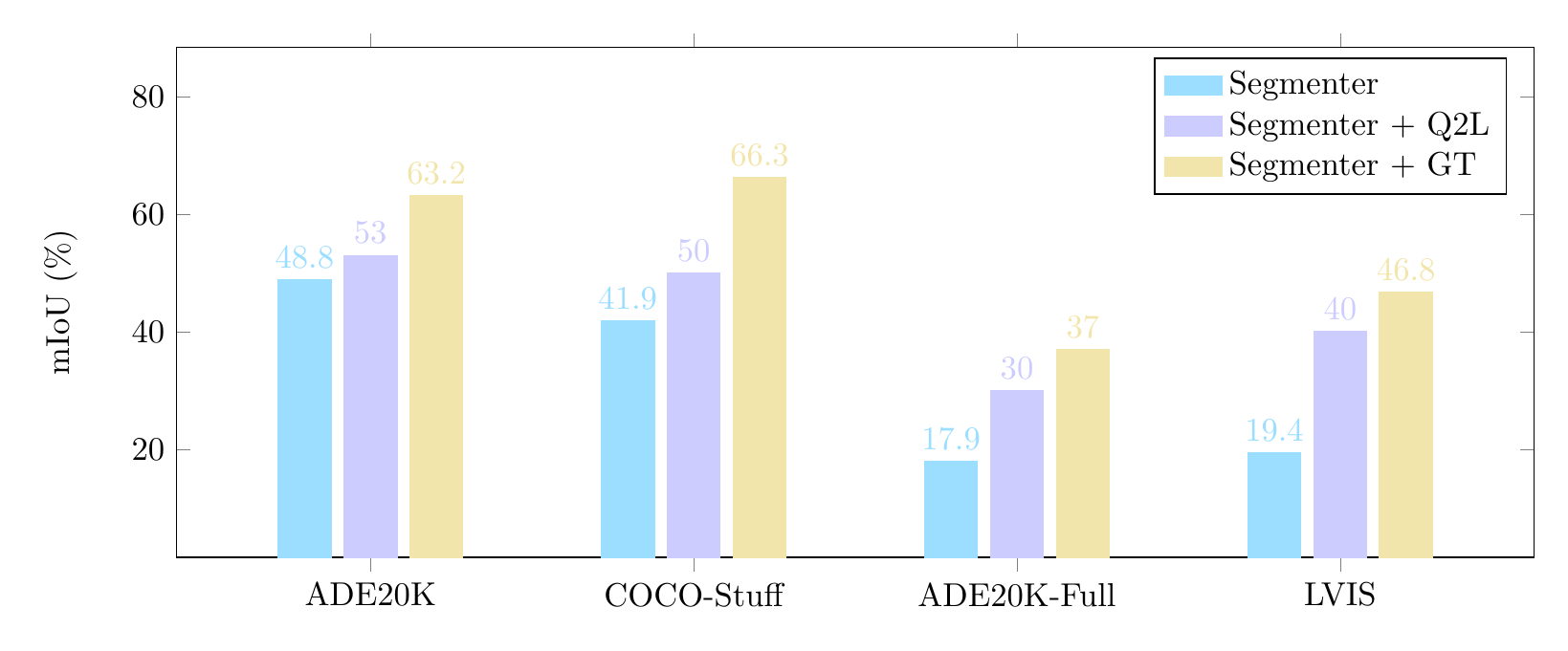}}
\caption{\tiny{Image semantic seg.}}
\end{subfigure}
\begin{subfigure}[b]{0.255\textwidth}
{\includegraphics[width=\textwidth,height=25mm]{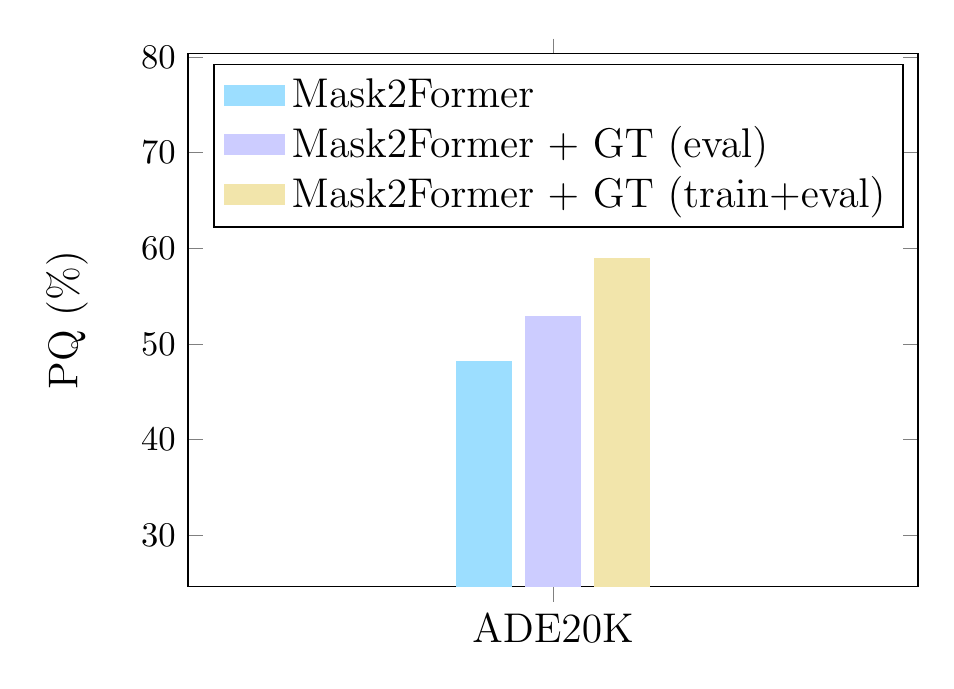}}
\caption{\tiny{Image panoptic seg.}}
\end{subfigure}
\begin{subfigure}[b]{0.255\textwidth}
{\includegraphics[width=\textwidth,height=25mm]{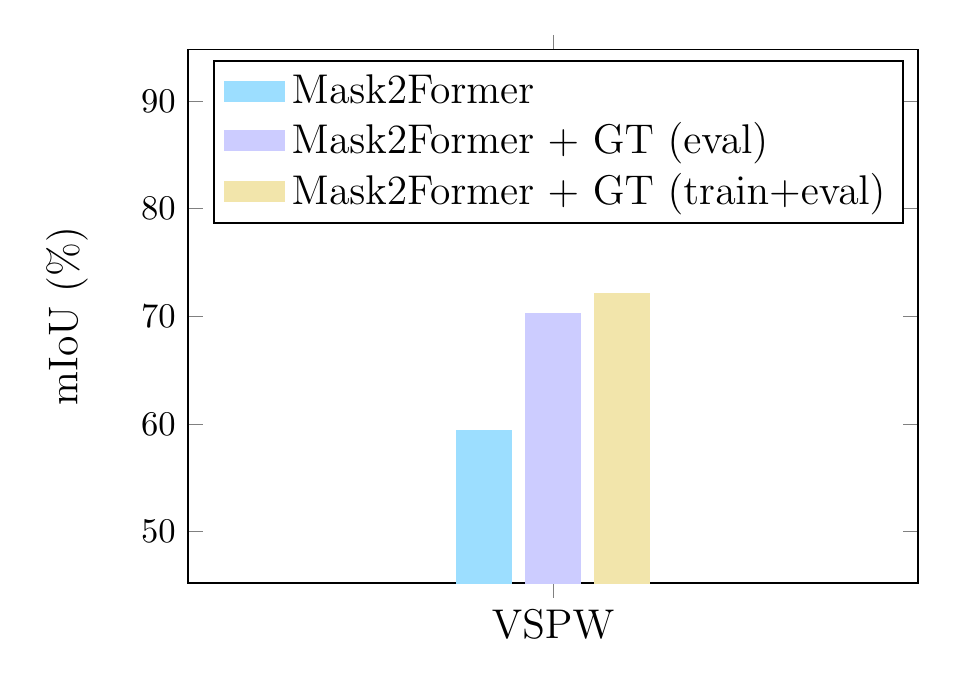}}
\caption{\tiny{Video semantic seg.}}
\end{subfigure}
\caption{
\small{\textbf{
Illustrating the effectiveness of multi-label classification:
} The segmentation results of exploiting the ground-truth multi-label during only evaluation or both training and evaluation on (a) PASCAL-Context, COCO-Stuff,
ADE$20$K-Full, and COCO+LVIS (image semantic segmentation)/(b) ADE$20$K (image panoptic segmentation)/(c) VSPW (video semantic segmentation). Refer to Sec.\ref{analysis} for more details. YouTubeVIS results are not included as we can not access the ground-truth.}}
\label{fig:gt_rankseg}
\end{minipage}
\end{figure*}

Motivated by the significant gains obtained with the ground-truth multi-label,
we propose two different schemes to exploit the benefit of multi-label image predictions including the \emph{independent single-task scheme} and \emph{joint multi-task scheme}.
For the independent single-task scheme, we train one model for multi-label image/video classification and another model for segmentation. Specifically, we first train a model to predict multi-label classification probabilities for each image/video, then we estimate the existing label subset for each image/video based on the multi-label predictions, last we use the predicted label subset to train the segmentation model for rank-adaptive selected-label pixel classification during both training and testing.
For the joint multi-task scheme, we train one model to support both multi-label image/video classification and segmentation based on a shared backbone.
Specifically, we apply a multi-label prediction head and a segmentation head, equipped with a rank-adaptive adjustment scheme, over the shared backbone and train them jointly.
For both schemes, we need to send the multi-label predictions
into the segmentation head for rank-adaptive selected-label pixel classification, which enables selecting a collection of categories that appear and adjusting the pixel-level classification scores according to the image/video content adaptively.

We demonstrate the effectiveness of our approach on various strong baseline methods
including DeepLabv$3$~\cite{chen2017rethinking}, Segmenter~\cite{strudel2021}, Swin-Transformer~\cite{liu2021swin}, \textsc{BEiT}~\cite{bao2021beit}, MaskFormer~\cite{cheng2021per}, Mask$2$Former~\cite{cheng2021masked}, and ViT-Adapter~\cite{chen2022vitadapter} across multiple segmentation benchmarks including PASCAL-Context~\cite{mottaghi2014role}, ADE$20$K~\cite{zhou2017scene}, COCO-Stuff~\cite{caesar2018coco}, ADE$20$K-Full~\cite{zhou2017scene,cheng2021per}, COCO+LVIS~\cite{gupta2019lvis,jain2021scaling}, YouTubeVIS~\cite{yang2019video}, and VSPW~\cite{miao2021vspw}.

\section{Related Work}

\noindent\textbf{Image segmentation}.
We can roughly categorize the existing studies
on image semantic segmentation into two main paths:
(i) region-wise classification methods~\cite{arbelaez2012semantic,caesar2016region,gu2009recognition,gould2009decomposing,wei2017object,neuhold2017mapillary,caesar2016region,uijlings2013selective},
which first organize the pixels into a set of regions (usually super-pixels),
and then classify each region to get the image segmentation result.
Several very recent methods~\cite{cheng2021per,zhang2021k,wang2021max}
exploit the DETR framework~\cite{carion2020end} to conduct region-wise classification
more effectively;
(ii) pixel-wise classification methods,
which predict the label of each pixel directly and dominate most previous studies
since the pioneering FCN~\cite{long2015fully}.
There exist extensive follow-up studies that improve the pixel
classification performance via constructing better contextual representations~\cite{zhao2017pyramid,chen2017rethinking,yuan2018ocnet,yuan2020object} or designing more effective decoder architectures~\cite{badrinarayanan2015segnet,ronneberger2015u,chen2018encoder}.
Image panoptic segmentation~\cite{kirillov2019panoptic,kirillov2019panoptic2} aims to unify image semantic segmentation and image instance segmentation tasks. Some recent efforts have introduced
various advanced architectures such as Panoptic FPN~\cite{kirillov2019panoptic2}, Panoptic DeepLab~\cite{cheng2019panoptic}, Panoptic Segformer~\cite{li2021panoptic}, K-Net~\cite{zhang2021k}, and Mask$2$Former~\cite{cheng2021masked}.
Our RankSeg is complementary with various paradigms and consistently improves several representative state-of-the-art methods across both image semantic segmentation and image panoptic segmentation tasks.

\noindent\textbf{Video segmentation}.
Most of the previous works address the video segmentation task by extending
the existing image segmentation models with temporal consistency constraint~\cite{wang2021survey}.
Video semantic segmentation aims to predict the semantic category of all pixels
in each frame of a video sequence,
where the main efforts focus on two paths including exploiting cross-frame relations to improve
the prediction accuracy~\cite{kundu2016feature,hur2016joint,jin2017video,gadde2017semantic,nilsson2018semantic,chandra2018deep} and leveraging the information of neighboring frames to accelerate computation~\cite{mahasseni2017budget,xu2018dynamic,li2018low,jain2019accel,hu2020temporally,liu2020efficient}.
Video instance segmentation~\cite{yang2019video} requires simultaneous detection, segmentation and tracking of
instances in videos and there exist four mainstream frameworks including tracking-by-detection~\cite{voigtlaender2019mots,lin2020video,cao2020sipmask,fu2020compfeat,hu2019learning}, clip-and-match~\cite{bertasius2020classifying,athar2020stem}, propose-and-reduce~\cite{lin2021video}, and segment-as-a-whole~\cite{wang2021end,hwang2021video,wu2021seqformer,cheng2021mask2former}.
We show the effectiveness of our method on both video semantic segmentation and video instance segmentation tasks via improving the very recent state-of-the-art method Mask$2$Former~\cite{cheng2021mask2former}.

\noindent\textbf{Multi-label classification}.
The goal of multi-label classification is to identity
all the categories presented in a given image or video over the complete label set.
The conventional multi-label image classification literature partitions the existing methods into three main directions:
(i) improving the multi-label classification loss functions to handle the imbalance issue~\cite{wu2020distribution,ben2020asymmetric},
(ii) exploiting the label co-occurrence (or correlations) to model the semantic relationships between different categories~\cite{hu2016learning,li2016conditional,chen2019learning,ye2020attention},
and (iii) localizing the diverse image regions associated with different categories~\cite{wang2017multi,guo2019visual,you2020cross,lanchantin2021general,liu2021query2label}.
In our independent single-task scheme,
we choose the very recent state-of-the-art method Query2Label~\cite{liu2021query2label}
to perform multi-label classification on various semantic segmentation benchmarks
as it is a very simple and effective method that exploits the benefits of both
label co-occurrence and localizing category-dependent regions.
There exist few efforts that apply multi-label image classification
to address segmentation task. To the best of our knowledge,
the most related study EncNet~\cite{zhang2018context} simply adds a multi-label
image classification loss w/o changing the original semantic segmentation head
that still needs to select the label of each pixel from all predefined categories.
We empirically show the advantage of our method over EncNet in the ablation experiments.
Besides, our proposed method is naturally suitable to solve large-scale
semantic segmentation problem as we only perform rank-adaptive pixel classification
over a small subset of the complete label set based on the
multi-label image prediction. We also empirically
verify the advantage of our method over the very recent ESSNet~\cite{jain2021scaling} in the ablation experiments.

\section{Our Approach}

We first introduce the overall framework of our approach in Sec.~\ref{framework}, which is also illustrated in Figure~\ref{fig:our_method}.
Second, we introduce the details of the independent single-task scheme in Sec.~\ref{single_task}
and those of joint multi-task scheme in Sec.~\ref{multi_task}.
Last, we conduct analysis experiments to
investigate the detailed improvements of our method across multiple segmentation tasks in Sec.~\ref{analysis}.

\subsection{Framework}\label{framework}
The overall framework of our method is illustrated in Figure~\ref{fig:our_method}, which consists of one path for multi-label image classification
and one path for semantic segmentation.
The multi-label prediction is used to sort and select the top $\kappa$
category embeddings with the highest confidence scores which are then sent into the rank-adaptive selected-label pixel classifier to generate the semantic segmentation prediction.
We explain the mathematical formulations of both
multi-label image classification
and rank-adaptive selected-label pixel classification as follows:

{
\definecolor{person}{rgb}{0, 0.8, 0}
\definecolor{grass}{rgb}{0, 0, 0.6}
\definecolor{railing}{rgb}{0.8, 0, 0}
\begin{figure*}[t]
\centering
\includegraphics[width=1\textwidth]{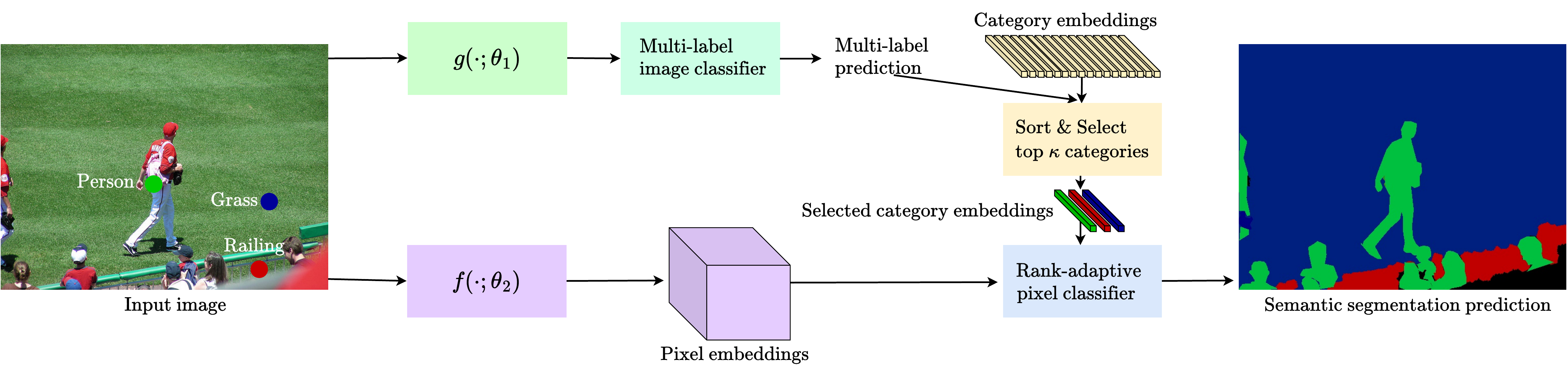}
\caption{\small{
\textbf{
Illustrating the framework of our approach:}
Given an input image that contains \textcolor{person}{person}, \textcolor{grass}{grass}, and \textcolor{railing}{railing},
which are marked with {\color{person}{\ding{108}}},~{\color{grass}{\ding{108}}},
and {\color{railing}{\ding{108}}}, respectively.
First, we use the multi-label image classification model $g(\cdot; \theta_1)$ and
multi-label image classifier to predict the presence probabilities of all categories.
Second, we sort the labels according to the predicted presence probabilities and select the
top $\kappa$ most probable category embeddings.
Last, we send the selected subset of category embeddings
into the rank-adaptive selected-label pixel classifier to identify
the label of each pixel based on the pixel embeddings output by the semantic segmentation model $f(\cdot; \theta_2)$.}}
\label{fig:our_method}
\end{figure*}
}

\noindent\textbf{Multi-label image classification}.
The goal of multi-label image classification is to predict the set of
existing labels in a given image $\mathbf{x} \in \mathbb{R}^{H\times W\times 3}$,
where $H$ and $W$ represent the input height and width.
We generate the image-level multi-label ground truth $\mathbf{y}^{gt}$ from the ground truth segmentation map and we represent $\mathbf{y}^{gt}$ with a vector of $K$ binary values $[{y}^{gt}_1,{y}^{gt}_2,...,{y}^{gt}_{K}]^\intercal, y^{gt}_i \in \{0,1\}$, where $K$ represents the total number of predefined categories and $y^{gt}_i=1$ represents the existence of pixels belonging to $i$-th category and $y^{gt}_i=0$ otherwise.

The prediction $\mathbf{y} \in \mathbb{R}^{K}$ is a vector that records the existence confidence score of each category in the given image $\mathbf{x}$.
We use $g(\cdot; \theta_1)$ to represent the backbone for the multi-label image classification model. We estimate the multi-label predictions of input $\mathbf{x}$ with the following sigmoid function:
\begin{align}
{y}_{k} = \frac{e^{\xi(g(\mathbf{x}; \theta_1), \mathbf{h}_{k})}}
{e^{\xi(g(\mathbf{x}; \theta_1), \mathbf{h}_{k})} + 1}, \label{eq:ml}
\end{align}
where ${y}_{k}$ is the $k$-th element of $\mathbf{y}$, $g(\mathbf{x}; \theta_1)$
represents the output feature map, $\mathbf{h}_{k}$ represents the multi-label image classification weight associated with the $k$-th category, and $\xi(\cdot)$ represents a transformation function that estimates the similarity between the output feature map and the multi-label image classification weights.
We supervise the multi-label predictions with the asymmetric loss that operates differently on positive and negative samples by following~\cite{ben2020asymmetric,liu2021query2label}.

\noindent\textbf{Rank-adaptive selected-label pixel classification}.
The goal of semantic segmentation is to predict the semantic label of each pixel
and the label is selected from all predefined categories.
We use $\mathbf{z} \in \mathbb{R}^{H\times W\times K}$ to represent
the predicted pixel classification probability map for the input image $\mathbf{x}$.
We use $f(\cdot; \theta_2)$ to represent the semantic segmentation backbone
and ${\mathbf{z}^{gt}} \in \mathbb{R}^{H\times W}$ to represent the ground-truth
segmentation map.
Instead of choosing the label of each pixel from all $K$ predefined categories,
based on the previous multi-label prediction $\mathbf{y}$ for image $\mathbf{x}$,
we introduce a more effective rank-adaptive selected-label pixel classification scheme:

\begin{itemize}
\item {Sort and select the top $\kappa$ elements of the classifier weights according to the descending order of multi-label predictions $\mathbf{y}=[y_1, y_2, \cdots, y_K]$}:
\begin{align}
[\overline{\mathbf{w}}_{1}, \overline{\mathbf{w}}_{2}, \cdots, \overline{\mathbf{w}}_{\kappa}] = \operatorname{Top-\kappa}([\mathbf{w}_{1}, \mathbf{w}_{2}, \cdots, \mathbf{w}_{K}], \mathbf{y}),
\end{align}
\item {Rank-adaptive classification of pixel $(i, j)$ over the top $\kappa$ selected categories}:
\begin{align}
\mathbf{z}_{i,j,k} = \frac{e^{\psi(f(\mathbf{x}; \theta_2)_{i,j}, \overline{\mathbf{w}}_{k})/{\tau_k}}}
{\sum_{l=1}^{\kappa} e^{{\psi(f(\mathbf{x}; \theta_2)_{i,j}, \overline{\mathbf{w}}_{l})}/{\tau_l}}}, \label{eq:eq3}
\end{align}
\end{itemize}
where $[\mathbf{w}_{1}, \mathbf{w}_{2}, \cdots, \mathbf{w}_{K}]$ represents the pixel classification weights for all $K$ predefined categories and $[\overline{\mathbf{w}}_{1}, \overline{\mathbf{w}}_{2}, \cdots, \overline{\mathbf{w}}_{\kappa}]$ represents the top $\kappa$ selected pixel classification weights associated with the largest multi-label classification scores.
$f(\mathbf{x}; \theta_2)$ represents the output feature map for semantic segmentation.
$\psi(\cdot)$ represents a transformation function that estimates the similarity between the pixel features and the pixel classification weights.
$\kappa$ represents the number of selected category embeddings and
$\kappa$ is chosen as a much smaller value than $K$.
{\begin{figure}[t]
\centering
\begin{subfigure}[b]{0.45\linewidth}
\centering
\includegraphics[width=1\textwidth]{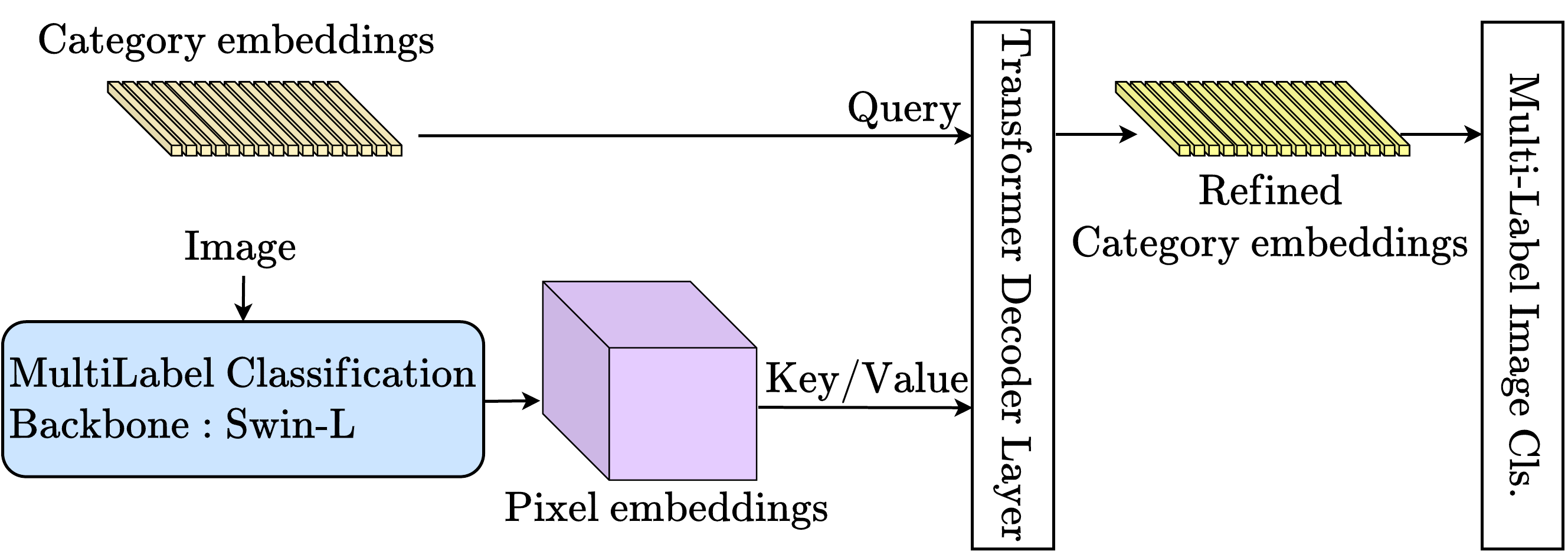}
\caption{\tiny{The framework of Query$2$Label}}
\label{fig:query2label}
\end{subfigure}%
\begin{subfigure}[b]{0.45\linewidth}
\centering
\includegraphics[width=1\textwidth]{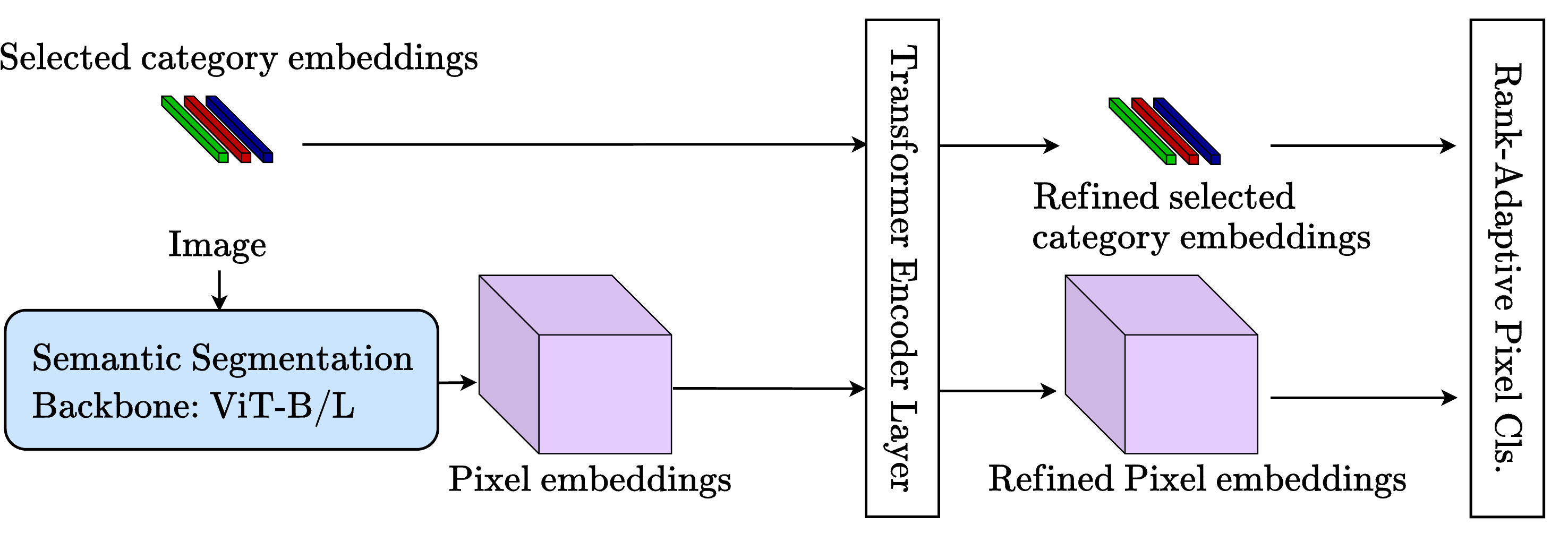}
\caption{\tiny{Segmenter w/ selected category embeddings}}
\label{fig:segmenter}
\end{subfigure}
\caption{\small
(a) \textbf{Illustrating the framework of Query$2$Label}:
The multi-label image classification backbone $g(\cdot; \theta_1)$ is set as Swin-L by default.
The transformation $\xi(\cdot)$ is implemented as two transformer decoder layers
followed by a linear layer that prepares the refined category embeddings
for the multi-label image classifier.
(b) \textbf{Illustrating the framework of Segmenter w/ selected category embeddings:} The semantic segmentation backbone $f(\cdot; \theta_2)$ is set as ViT-B/$16$ or ViT-L/$16$. The transformation $\psi(\cdot)$ is implemented as two transformer encoder layers followed by $\ell_2$-normalization before estimating the segmentation map.
}
\end{figure}}
We apply a set of rank-adaptive learnable temperature parameters $[{\tau}_{1}, {\tau}_{2}, \cdots, {\tau}_{\kappa}]$ to adjust the classification scores over the selected top $\kappa$ categories.
The temperature parameters across different selected classes are shared in all of the baseline experiments by default\footnote{We set $\tau_1$=$\tau_2$=$\cdots$=$\tau_\kappa$ for all baseline segmentation experiments.}.
We analyze the influence of $\kappa$ choices and the benefits of such a rank-oriented adjustment scheme in the following discussions and experiments.

\subsection{Independent single-task scheme}\label{single_task}
Under the independent single-task setting,
the multi-label image classification model $g(\cdot; \theta_1)$ and
the semantic segmentation model $f(\cdot; \theta_2)$ are
trained separately and their model parameters are not shared, i.e., $\theta_1\neq \theta_2$.
Specifically, we first train the multi-label image classification model $g(\cdot; \theta_1)$
to identify the top $\kappa$ most likely categories for each image.
Then we train the rank-adaptive selected-label pixel classification
model, i.e., semantic segmentation model, $f(\cdot; \theta_2)$
to predict the label of each pixel over the selected top $\kappa$ classes.

\noindent\textbf{Multi-label image classification model}.
We choose the very recent SOTA multi-label classification method Query$2$Label~\cite{liu2021query2label} as it performs best on multiple multi-label classification benchmarks by the time of our submission according to {paper-with-code}.\footnote{https://paperswithcode.com/task/multi-label-classification}
The key idea of Query$2$Label is to use the category embeddings
as the query to gather the desired pixel embeddings as the key/value, which is output by an ImageNet-$22$K pre-trained backbone such as Swin-L, adaptively
with one or two transformer decoder layers.
Then Query$2$Label scheme applies a multi-label image classifier over the refined category embeddings to predict the existence of each category.
Figure~\ref{fig:query2label} illustrates the framework of Query$2$Label framework.
Refer to~\cite{liu2021query2label}
and the official implementation for more details.
The trained weights of Query$2$Label model are fixed
during both training and inference of the following semantic segmentation model.

{
\begin{figure}[t]
\centering
\includegraphics[width=1\textwidth]{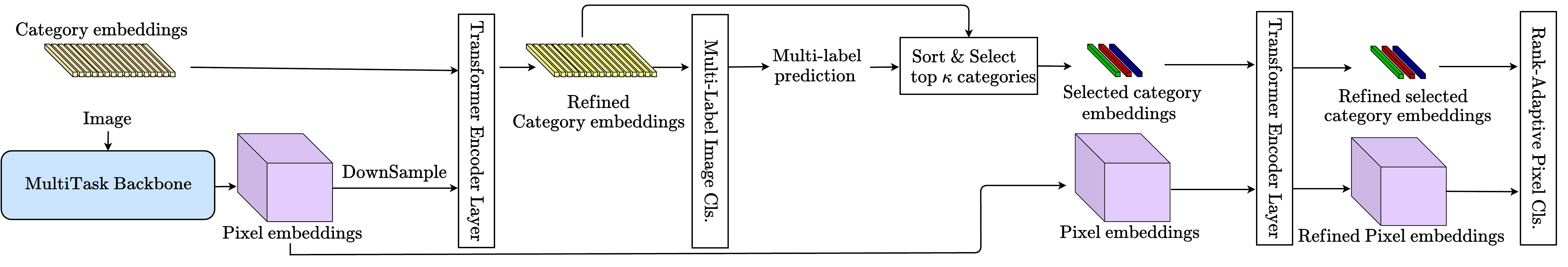}
\caption{\small
\textbf{
Illustrating the framework of joint multi-task scheme:}
$g(\cdot; \theta_1)$ and $f(\cdot; \theta_2)$ are set as the shared multi-task
backbone and $\theta_1=\theta_2$.
$\xi(\cdot)$ is implemented as one transformer encoder layer or two transformer decoder layers or global average pooling + linear projection.
$\psi(\cdot)$ is implemented as two transformer encoder layers
followed by L$2$-normalization before estimating the segmentation map.
}
\label{fig:multitask}

\end{figure}
}

\noindent\textbf{Rank-adaptive selected-label pixel classification model}.
We choose a simple yet effective baseline Segmenter~\cite{strudel2021} as it achieves even better performance than Swin-L\footnote{Segmenter w/ ViT-L: $53.63\%$ vs. Swin-L: $53.5\%$ on ADE$20$K.}
when equipped with ViT-L.
Segmenter first concatenates the category embedding with the pixel embeddings output
by a ViT model together and then sends them into two transformer encoder layers.
Last, based on the refined pixel embeddings and category embeddings, Segmenter computes
their $\ell_2$-normalized scalar product as the segmentation predictions.
We select the top $\kappa$ most likely categories for each image according to the predictions of the Query$2$Label model and only use the selected top $\kappa$ category embeddings instead of all category embeddings.
Figure~\ref{fig:segmenter} illustrates the overall framework of Segmenter with the selected category embeddings.

\subsection{Joint multi-task scheme}\label{multi_task}
Considering that the independent single-task scheme suffers
from extra heavy computation overhead as the
Query$2$Label method relies on a large backbone, e.g., Swin-L,
we introduce a joint multi-task scheme
that shares the backbone for both sub-tasks, in other words,
$\theta_1 = \theta_2$ and the computations of
$g(\cdot; \theta_1)$ and $f(\cdot; \theta_2)$ are also shared.

Figure~\ref{fig:multitask} shows the overall framework
of the joint multi-task scheme.
First, we apply a shared multi-task backbone to process the input image
and output the pixel embeddings.
Second, we concatenate the category embeddings
with the down-sampled pixel embeddings\footnote{Different from the semantic segmentation task, the multi-label image classification task does not require high-resolution representations.},
send them into one transformer encoder layer, and apply the multi-label image
classifier on the refined category embeddings to
estimate the multi-label predictions.
Last, we sort and select the top $\kappa$ category embeddings,
concatenate the selected category embeddings with the pixel embeddings,
send them into two transformer encoder layers,
and compute the semantic segmentation predictions
based on $\ell_2$-normalized scalar product between the refined selected
category embeddings and the refined pixel embeddings.
We empirically verify the advantage of the joint multi-task scheme over the independent single-task scheme in the ablation experiments.

\subsection{Analysis experiments}\label{analysis}

\noindent\textbf{Oracle experiments.}
We first conduct several groups of oracle experiments based on Segmenter w/ ViT-B/$16$
on four challenging image semantic segmentation benchmarks ( PASCAL-Context/COCO-Stuff/ADE$20$K-Full/COCO+LVIS), Mask$2$Former w/ Swin-L on both ADE$20$K panoptic segmentation benchmark\footnote{We choose Swin-L by following the MODEL\_ZOO of the official Mask$2$Former implementation: https://github.com/facebookresearch/Mask2Former} and VSPW video semantic segmentation benchmark.

\noindent\emph{- Segmenter/Mask$2$Former + GT (train + eval)}: the upper-bound segmentation performance of Segmenter/Mask$2$Former when training \& evaluating equipped with the ground-truth multi-label of each image or video,
in other words, we only need to select the category of each pixel over the
ground-truth existing categories in a given image or video.

\noindent\emph{- Segmenter/Mask$2$Former + GT (eval)}: the upper-bound segmentation performance of Segmenter/Mask$2$Former when only using the ground-truth multi-label of each image or video during evaluation.

Figure~\ref{fig:gt_rankseg} illustrates the detailed comparison results. We can see that only applying the ground-truth multi-label
during evaluation already brings considerable improvements and
further applying the ground-truth multi-label during training
significantly improves the segmentation performance across all benchmarks.
For example, when compared to the baseline Segmenter or Mask$2$Former,
Segmenter + GT (train + eval) gains +$17\%$/+$24\%$/+$19\%$/+$27\%$ absolute
mIoU scores across PASCAL-Context/COCO-Stuff/ADE$20$K-Full/COCO+LVIS and
Mask$2$Former + GT (train + eval)
gains +$11\%$/+$13\%$ on ADE$20$K/VSPW.

\noindent\textbf{Improvement analysis of RankSeg.}
Table~\ref{tab:gtstudy} reports the results with different combinations of the proposed components within our joint multi-task scheme.
We can see that:
(i) multi-task learning (MT) introduces the auxiliary multi-label image classification task and brings relatively minor gains on most benchmarks,
(ii) combining MT with label sorting \& selection (LS) achieves considerable gains,
and
(iii) applying the rank-adaptive-$\tau$ manner (shown in Equation~\ref{eq:eq3}) instead of shared-$\tau$ achieves better performance.
We investigate the possible reasons by analyzing the value distribution of learned $1/\tau$ with the rank-adaptive-$\tau$ manner in Figure~\ref{fig:tau},
which shows that the learned $1/\tau$ is capable of adjusting the pixel classification scores based on the order of multi-label classification scores.
In summary, we choose ``MT + LS + RA'' scheme by default, which gains
+$0.91\%$/+$3.13\%$/+$0.85\%$/+$1.85\%$/+$0.8\%$/+$0.7\%$ over the baseline methods across these six challenging segmentation benchmarks respectively.

\begin{table}[t]
\centering
\setlength{\tabcolsep}{0.2pt}
\footnotesize
\renewcommand{\arraystretch}{1.1}
\caption{\small Ablation of the improvements with our method.
MT: multi-task learning with auxiliary multi-label image classification scheme.
LS: label sort and selection, in other words, sort and select the top $\kappa$ classes.
RA: rank-adaptive-$\tau$, which applies independent $\tau$ for pixel classification scores associated with the different ranking positions.
}
\label{tab:gtstudy}
\resizebox{\linewidth}{!}
{
\begin{tabular}{ l | c | c | c | c | c | c}
\shline
&  \multicolumn{4}{c|}{Image semantic seg.} & \multicolumn{1}{c|}{Image panoptic seg.} & \multicolumn{1}{c}{Video semantic seg.}\\
\cline{2-7}
Method.  &  PASCAL-Context & COCO-Stuff & ADE$20$K-Full & COCO+LVIS & ADE$20$K & VSPW \\
\shline
Baseline    &   $53.85$     &   $41.85$    &   $17.93$      &    $19.41$ &  $48.1$ & $59.4$ \\
+ MT        &   $54.05$     &   $42.38$     &    $17.81$    &  $20.26$    &  $48.2$ &  $59.5$  \\
+ MT + LS      &   $54.27$     &   $44.31$     &    $18.26$    &  $21.13$    &   $48.8$ & $59.6$  \\
+ MT + LS + RA  &   $\bf54.76$     &   $\bf44.98$    &   $\bf18.78$      &    $\bf21.26$   & $\bf48.9$ & $\bf60.1$ \\
\shline
\end{tabular}
}
\end{table}

{
\begin{figure}[htb]
\centering
\begin{subfigure}[b]{0.245\linewidth}
\centering
\includegraphics[width=1\textwidth]{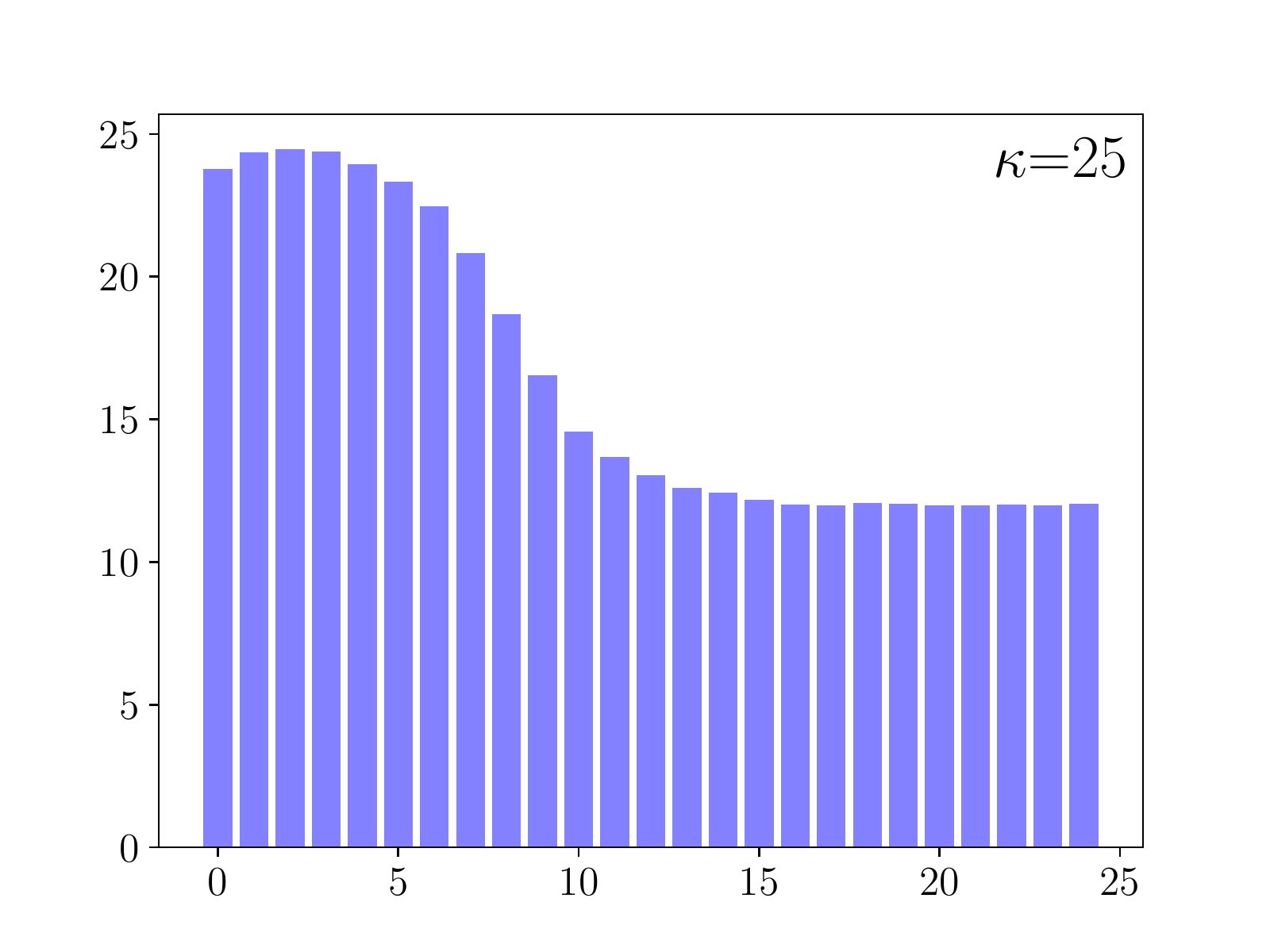}
\caption{\tiny{PASCAL-Context}}
\label{fig:gamma_ade}
\end{subfigure}%
\begin{subfigure}[b]{0.245\linewidth}
\centering
\includegraphics[width=1\textwidth]{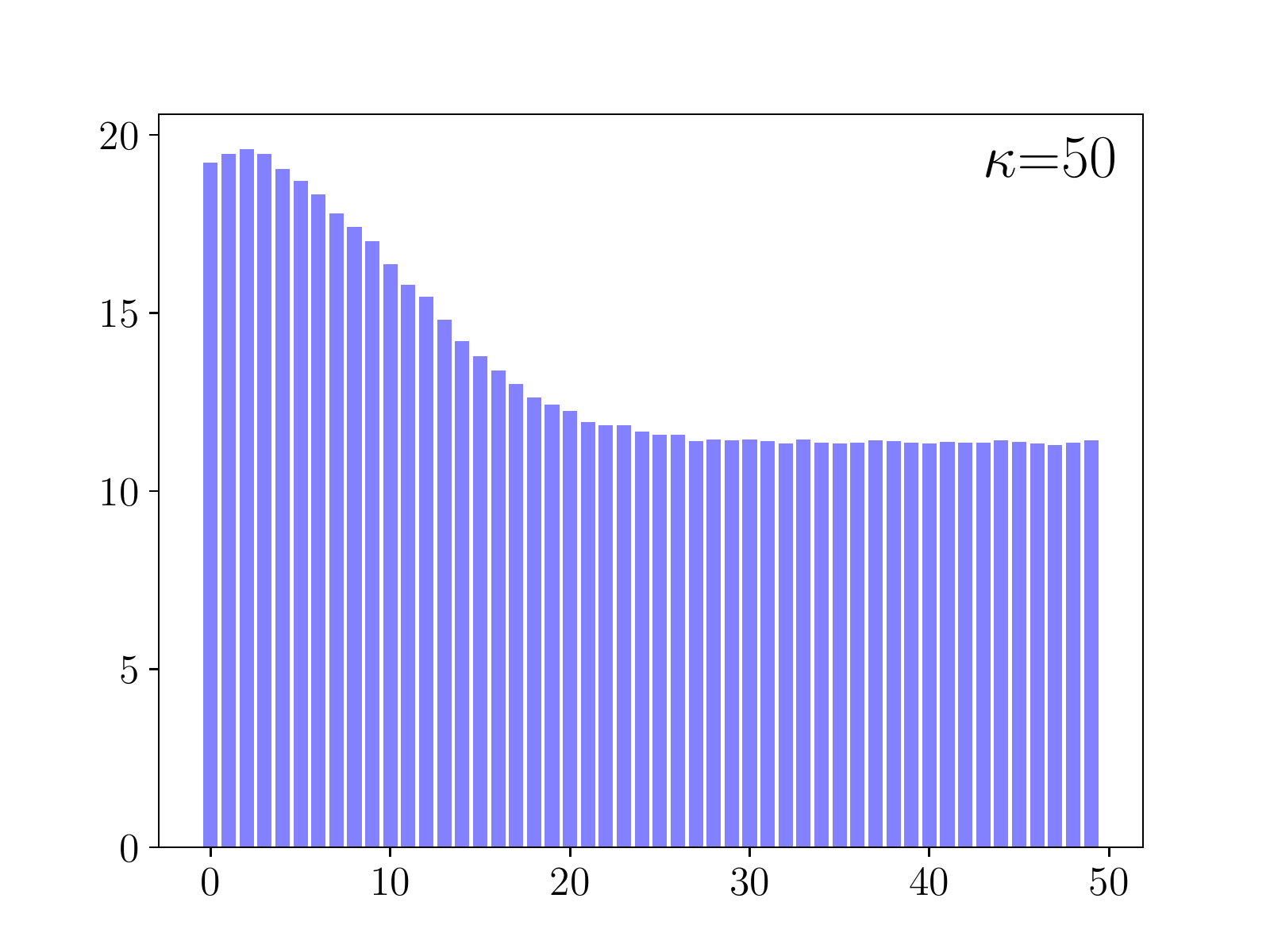}
\caption{\tiny{COCO-Stuff}}
\label{fig:gamma_cocostuff}
\end{subfigure}
\begin{subfigure}[b]{0.245\linewidth}
\centering
\includegraphics[width=1\textwidth]{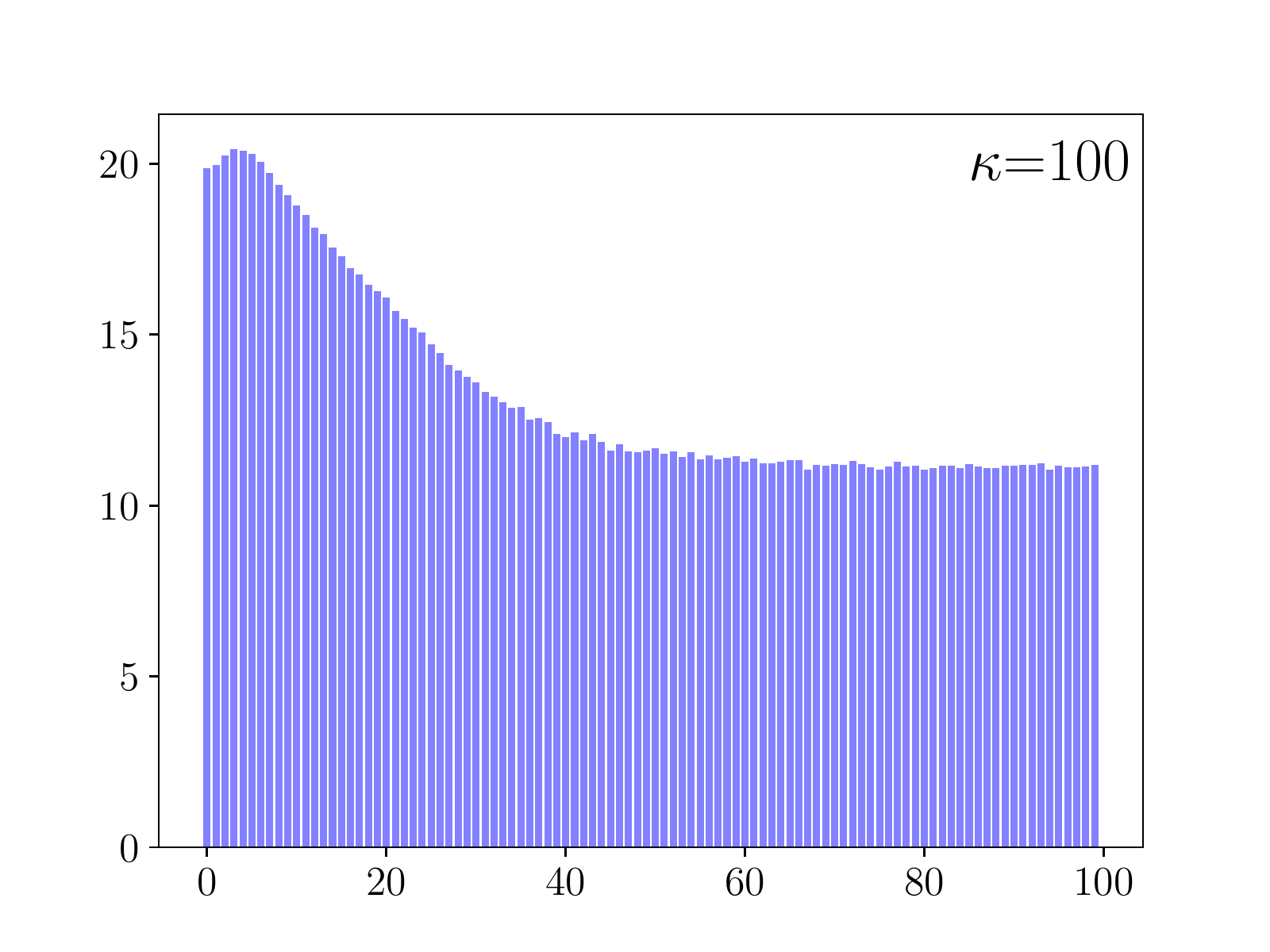}
\caption{\tiny{ADE$20$K-Full}}
\label{fig:gamma_cocostuff}
\end{subfigure}
\begin{subfigure}[b]{0.245\linewidth}
\centering
\includegraphics[width=1\textwidth]{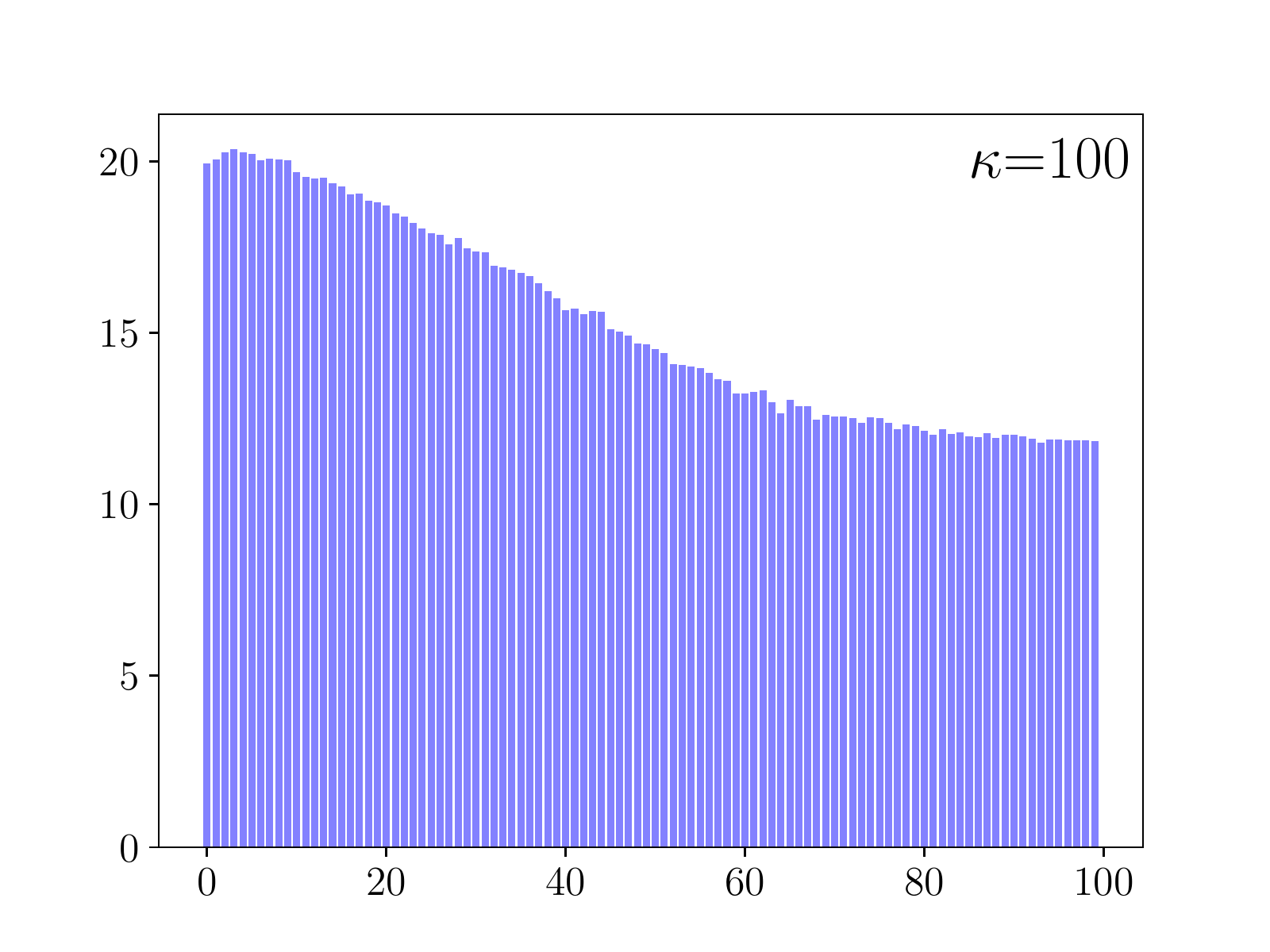}
\caption{\tiny{COCO+LVIS}}
\label{fig:gamma_cocostuff}
\end{subfigure}
\caption{\small
Illustrating the values of learned $1/\tau$ with rank-adaptive-$\tau$ manner on four semantic segmentation benchmarks.
We can see that the values of $1/\tau$ are almost monotonically decreasing, thus meaning that the pixel classification scores of the category associated with larger multi-label classification scores are explicitly increased.
}
\label{fig:tau}
\end{figure}
}

\section{Experiment}

\subsection{Implementation details}
We illustrate the details of the datasets, including ADE$20$K~\cite{zhou2017scene}, ADE$20$K-Full~\cite{zhou2017scene,cheng2021per}, PASCAL-Context~\cite{mottaghi2014role}, COCO-Stuff~\cite{caesar2018coco}, COCO+LVIS~\cite{gupta2019lvis,jain2021scaling}, VSPW~\cite{miao2021vspw}, and YouTubeVIS~\cite{yang2019video}, in the supplementary material.

\noindent\textbf{Multi-label image classification}.
For the independent single-task scheme, following the official implementation\footnote{https://github.com/SlongLiu/query2labels}
of Query$2$Label, we train multi-label image classification models, e.g., ResNet-$101$~\cite{he2016deep}, TResNetL~\cite{ridnik2021imagenet21k}, and Swin-L~\cite{liu2021swin}, for $80$ epochs using Adam solver with early stopping.
Various advanced tricks such as cutout~\cite{devries2017improved}, RandAug~\cite{cubuk2020randaugment} and EMA~\cite{he2020momentum} are also used.
For the joint multi-task scheme, we simply train the multi-label image classification
models following the same settings as the segmentation models w/o using the above-advanced tricks that might influence the segmentation performance.
We illustrate more details of the joint multi-task scheme in the supplementary material.

\noindent\textbf{Segmentation}.
We adopt the same settings for both independent single-task scheme
and joint multi-task scheme.
For the segmentation experiments based on Segmenter~\cite{strudel2021}, DeepLabv$3$~\cite{chen2017rethinking}, Swin-Transformer~\cite{liu2021swin}, \textsc{BEiT}~\cite{bao2021beit}, and ViT-Adapter~\cite{chen2022vitadapter}.
we follow the default training \& testing settings of their reproduced
version based on \texttt{mmsegmentation}~\cite{mmseg2020}.
For the segmentation experiments based on MaskFormer or Mask$2$Former, we follow their official implementation\footnote{https://github.com/facebookresearch/Mask2Former}.

\noindent\textbf{Hyper-parameters}.
We set $\kappa$ as $25$, $50$, $50$, $100$, and $100$ on PASCAL-Context, ADE$20$K, COCO-Stuff, ADE$20$K-Full, and COCO+LVIS respectively as they consist of a different number of semantic categories.
We set their multi-label image classification loss weights as $5$, $10$, $10$, $100$, and $300$. The segmentation loss weight is set as $1$.
We illustrate the hyper-parameter settings of experiments on MaskFormer~\cite{cheng2021per} or Mask$2$Former~\cite{cheng2021masked} in the supplementary material.

\noindent\textbf{Metrics}.
We report mean average precision (mAP) for multi-label image classification task, mean intersection over union (mIoU) for image/video semantic segmentation task, panoptic quality (PQ) for panoptic segmentation task,
and mask average precision (AP) for instance segmentation task.

\begin{table}[t]
\begin{minipage}[t]{1\linewidth}

\centering
\setlength{\tabcolsep}{0.1pt}
\scriptsize
\caption{\footnotesize Influence of the multi-label image classification accuracy (mAP) on semantic segmentation accuracy (mIoU) based on independent single-task manner.}
\resizebox{1\linewidth}{!}
{
\begin{tabular}{ l | c | c | c | c | c | c}
\shline
\multirow{2}{*}{Backbone} & \multicolumn{3}{c|}{COCO-Stuff} & \multicolumn{3}{c}{COCO+LVIS} \\
\cline{2-7}
 &  ResNet$101$~\cite{ridnik2021tresnet} &  TResNetL~\cite{ridnik2021tresnet} & Swin-L~\cite{liu2021swin}   &  ResNet$101$~\cite{ridnik2021tresnet} &  TResNetL~\cite{ridnik2021tresnet} &  Swin-L~\cite{liu2021swin}  \\ \hline
mAP ($\%$)      &   $55.17$    &  $60.10$   &   $\bf{64.79}$  & $26.54$  & $31.01$ & $\bf34.93$  \\ \hline
mIoU ($\%$)     &   $39.34$    &  $42.87$   &   $\bf{44.42}$ & $16.30$ & $19.88$ & $\bf21.19$   \\
$\bigtriangleup$    &   -$2.51$   &  +$1.02$  &  +$\bf2.57$  & -$3.11$ & +$0.47$ &   +$\bf1.78$  \\ \shline
\end{tabular}
}

\label{table:ablation_q2l_backbone}
\end{minipage}
\begin{minipage}[t]{1\linewidth}
\centering
\setlength{\tabcolsep}{1pt}
\footnotesize
\renewcommand{\arraystretch}{1.1}
\scriptsize
\caption{\footnotesize Independent single-task scheme vs. Joint multi-task scheme:
we adopt Swin-L as the backbone for the multi-label predictions in the independent single-task scheme.}
\resizebox{1\linewidth}{!}
{
\begin{tabular}{ l | c | c | c  c | c | c | c c}
\shline
\multirow{2}{*}{Method}  & \multicolumn{4}{c|}{COCO-Stuff} & \multicolumn{4}{c}{COCO+LVIS} \\
\cline{2-9}
&  {\#params.} & {FLOPs}  &  mIoU ($\%$) & $\bigtriangleup$ & {\#params.} & {FLOPs}  &  mIoU ($\%$) & $\bigtriangleup$ \\
\shline
Indep. single-task. &  $343.21$M & $182.8$G  &   $44.42$ & +$2.52$  & $347.33$M & $233.62$G & $21.19$ & +$1.78$\\
Joint multi-task.     & $\bf{109.73}$M  & $\bf{78.71}$G & $\bf{44.98}$ & +$\bf3.08$ & $\bf111.45$M & $\bf99.59$G & $\bf{21.26}$ & +$\bf1.85$\\
\shline
\end{tabular}}

\label{table:gtstudy_single_vs_multiple}
\end{minipage}
\begin{minipage}[t]{1\linewidth}
\centering
\setlength{\tabcolsep}{1pt}
\scriptsize
\caption{\small Influence of the size of the selected label set, i.e., $\kappa$.}
\resizebox{\linewidth}{!}
{
\begin{tabular}{ l | c | c | c | c | c | c | c | c | c | c | c}
\shline
\multirow{2}{*}{$\kappa$} & \multicolumn{7}{c|}{COCO-Stuff} & \multicolumn{4}{c}{COCO+LVIS} \\
\cline{2-12}
    &  $25$  &  $50$  &  $75$ &  $100$  &  $125$ & $150$ & $171$  &  $50$  &  $100$  &  $200$  & $1,284$ \\ \hline
mIoU ($\%$)     &   $44.65$    &  $\bf{44.98}$   &   $44.58$  &   $44.67$ &   $44.41$&   $44.37$ &   $44.20$   & $20.36$    &  $\bf{21.26}$   &   $20.99$   & $20.93$  \\ 
$\bigtriangleup$   &  +$2.80$   &  +$\bf3.13$  &  +$2.73$ &  +$2.82$ &  +$2.56$ &  +$2.52$   & +$2.35$ & +$0.95$ &  +$\bf{1.85}$ &  +$1.58$ & +$1.52$     \\
\shline
\end{tabular}}
\label{tab:ablation_labelset_size_1}

\end{minipage}
\begin{minipage}[t]{0.42\linewidth}
\centering
\footnotesize
\setlength{\tabcolsep}{0.5pt}
\caption{\small Influence of the multi-label classification loss weight.}
\resizebox{\linewidth}{!}
{
\begin{tabular}{ l | c | c | c | c }
\shline
\pbox{22mm}{multi-label\\cls. loss weight }   &  $1$   &  $5$  & $10$  &  $20$ \\ \hline
mAP ($\%$)           &  $59.14$ &   $62.38$    &   $\bf{62.52}$   &   $62.18$  \\\hline
mIoU ($\%$)          & $44.11$ &   $44.90$    &   $\bf{44.98}$   &   $44.04$ \\
$\bigtriangleup$   & +$2.26$ &   +$3.05$    &   +$\bf{3.13}$   &   +$2.19$ \\
\shline
\end{tabular}}

\label{tab:ablation_loss_weight}
\end{minipage}
\begin{minipage}[t]{0.55\linewidth}
\centering
\footnotesize
\setlength{\tabcolsep}{0.5pt}
\renewcommand{\arraystretch}{1.25}
\caption{\small Influence of multi-label prediction head architecture.}
\resizebox{\linewidth}{!}
{
\begin{tabular}{ l | c | c | c | c c }
\shline
Method             &  \#params. & FLOPs &  mAP & mIoU ($\%$) & $\bigtriangleup$ \\ \hline
GAP+Linear         &  $\bf{103.24}$M    &  $\bf{76.94}$G   &  $60.62$ &  $43.19$ & +$1.34$\\
$2\times$ TranDec  &   $114.27$M    &  $78.83$G   &  $61.15$ &  $44.07$ & +$2.22$\\
$1\times$ TranEnc  &   $109.73$M    &  $78.71$G    &  $\bf{62.52}$ &  $\bf{44.98}$ & +$\bf3.13$ \\
\shline
\end{tabular}}
\label{tab:ablation_mt_head}

\end{minipage}
\end{table}

\subsection{Ablation experiments}
We conduct all ablation experiments based on the Segmenter w/ ViT-B
and report their single-scale evaluation results on COCO-Stuff \texttt{test}
and COCO+LVIS \texttt{test} if not specified.
The baseline with Segmenter w/ ViT-B achieves $41.85\%$ and $19.41\%$ mIoU on COCO-Stuff and COCO+LVIS, respectively.

\noindent\textbf{Influence of the multi-label classification accuracy}.
We investigate the influence of multi-label classification accuracy on semantic segmentation tasks based on the independent single-task scheme.
We train multiple Query2Label models based on different backbones, e.g., ResNet$101$, TResNetL, and Swin-L.
Then we train three Segmenter w/ ViT-B segmentation models based on their multi-label predictions independently.
According to the results in Table~\ref{table:ablation_q2l_backbone},
we can see that more accurate multi-label classification prediction
brings more semantic segmentation performance gains and less accurate multi-label classification prediction even results in worse results than baseline.

\noindent\textbf{Independent single-task scheme vs. Joint multi-task scheme}.
We compare the performance and model complexity of the independent single-task scheme (w/ Swin-L) and joint multi-task scheme in Table~\ref{table:gtstudy_single_vs_multiple}.
To ensure fairness, we choose the \# of labels in the selected label set, i.e., $\kappa$, as $50$ for both schemes, and the segmentation models are trained \& tested under
the same settings.
According to the comparison results, we can see that joint multi-task scheme
achieves better performance with fewer parameters and FLOPs.
Thus, we choose the joint multi-task scheme in the following
experiments for efficiency if not specified.

\noindent\textbf{Influence of different top $\kappa$}.
We study the influence of the size of selected label set, i.e., $\kappa$, as shown in Table~\ref{tab:ablation_labelset_size_1}.
According to the results, our method achieves the best performance on COCO-Stuff/COCO+LVIS
when $\kappa$=$50$/$\kappa$=$100$, which achieves a better trade-off between precision and recall
for multi-label predictions.
Besides, we attempt to fix $\kappa$=$50$ during
training and report the results when changing $\kappa$ during evaluation on COCO-Stuff:
$\kappa$=$25$: $44.89\%$ /$\kappa$=$50$: $44.98\%$ /$\kappa$=$75$: $45.00\%$ /$\kappa$=$100$: $45.01\%$.
Notably, we also report the results with $\kappa$=$K$, in other words, we only sort the classifier weights, which also achieve considerable gains. Therefore, we can see that sorting the classes and rank-adaptive adjustment according to multi-label predictions are the key to the gains.
In summary, our method consistently outperforms baseline with
different $\kappa$ values.
We also attempt to use dynamic $\kappa$ for different images during evaluation
but observe no significant gains. More details are provided in the supplementary material.

\noindent\textbf{Influence of the multi-label classification loss weight}.
We study the influence of the multi-label image classification loss weights with the joint multi-task scheme on COCO-Stuff and report the results in Table~\ref{tab:ablation_loss_weight}.
We can see that setting the multi-label classification loss weight as $10$ achieves the best performance.

\noindent\textbf{Influence of $\xi(\cdot)$ choice}.
Table~\ref{tab:ablation_mt_head} compares the results based on different multi-label prediction head architecture choices including
``GAP+Linear'' (applying global average pooling followed by linear projection), ``$2\times$ TranDec'' (using two transformer decoder layers),
and ``$1\times$ TranEnc'' (using one transformer encoder layer) on COCO-Stuff.
Both ``$2\times$ TranDec'' and ``$1\times$ TranEnc'' operate on feature maps with
$\frac{1}{32}$ resolution of the input image.
According to the results, we can see that ``$1\times$ TranEnc'' achieves the best performance and we implement $\xi(\cdot)$ as one transformer encoder layer
if not specified.
We also attempt generating multi-label prediction (mAP=$58.55\%$) from the semantic segmentation prediction directly but observe no performance gains, thus verifying the importance of relatively more accurate multi-label classification predictions.

More comparison results with the previous EncNet~\cite{zhang2018context} and ESSNet~\cite{jain2021scaling} are summarized in the supplementary material.

\subsection{State-of-the-art experiments}

\noindent\textbf{Image semantic segmentation}.
We apply our method to various state-of-the-art image semantic segmentation methods including DeepLabv$3$, Seg-Mask-L/$16$, Swin-Transformer, \textsc{BEiT}, and ViT-Adapter-L.
Table~\ref{tab:sota_seg} summarizes the detailed comparison results across
three semantic segmentation benchmarks including ADE$20$K, COCO-Stuff, and COCO+LVIS, where we evaluate the multi-scale segmentation results on ADE$20$K/COCO-Stuff and single-scale segmentation results on COCO+LVIS (due to limited GPU memory) respectively. More details of how to apply our joint multi-task scheme to these methods are provided in the supplementary material.
According to the results in Table~\ref{tab:sota_seg},
we can see that our RankSeg consistently improves DeepLabv$3$, Seg-Mask-L/$16$, Swin-Transformer, \textsc{BEiT}, and ViT-Adapter-L across three
evaluated benchmarks.
For example, with our RankSeg, \textsc{BEiT} gains $0.8\%$ on ADE$20$K with slightly more parameters and GFLOPs.

\begin{table}[h]
\begin{minipage}[t]{1\linewidth}
\centering
\renewcommand{\arraystretch}{1.1}
\caption{\small Combination with DeepLabv$3$, Seg-Mask-L, Swin-B, \textsc{BEiT}, and ViT-Adapter.}
\resizebox{1\linewidth}{!}
{
\begin{tabular}{l|c|c|c|c|c|c|c|c|c}
\shline
\multirow{2}{*}{Method}  & \multicolumn{3}{c|}{ADE$20$K}  & \multicolumn{3}{c|}{COCO-Stuff} & \multicolumn{3}{c}{COCO+LVIS}\\
\cline{2-10}
& \#params. & FLOPs  & \pbox{30mm}{mIoU($\%$)} & \#params. & FLOPs  & mIoU($\%$) & \#params. & FLOPs  & mIoU($\%$) \\\shline
DeepLabv$3$          &  $\bf{87.21}$M  & $\bf{347.64}$G & $45.19$&  $\bf{87.22}$M  & $\bf{347.68}$G &  $38.42$  &  $\bf{88.08}$M  & $\bf{350.02}$G  & $11.04$   \\
\RANKSEG  &  $91.87$M   &  $349.17$G & $\bf{46.61}$ &  $91.75$M   &  $349.09$G &   $\bf{39.86}$   &  $93.21$M   &  $359.47$G &  $\bf{12.76}$  \\\hline
Seg-Mask-L/$16$   &  $\bf{333.23}$M & $377.83$G   & $53.63$ &  $\bf{333.26}$M & $378.57$G  &   $47.12$ &  $\bf{334.4}$M & $\bf{420.2}$G &  $23.71$  \\
\RANKSEG   &  $345.99$M  & $\bf{377.18}$G   & $\bf{54.47}$  &  $346.03$M  & $\bf{377.46}$G   &   $\bf{47.93}$ &  $348.31$M  & $422.6$G  &  $\bf{24.60}$  \\\hline
Swin-B   &  $\bf{121.42}$M  &  $\bf{299.81}$G   & $52.4$   &  $\bf{121.34}$M  &  $\bf{299.98}$G  &   $47.16$   &  $\bf{122.29}$M  &  $309.34$G  &  $20.33$ \\
\RANKSEG &  $125.43$M  &  $300.48$G & ${\bf53.01}$  &  $125.46$M  &  $300.56$G  &   $\bf{47.85}$ &  $126.89$M  &  $\bf{306.53}$G &  $\bf{20.81}$ \\\hline
\textsc{BEiT}    &  $\bf{441.27}$M  &  $\bf{1745.99}$G & $57.0$  &  $\bf{441.30}$M  &  $\bf{1746.54}$G  & $49.9$  &  \multicolumn{3}{c}{OOM}  \\
\RANKSEG &  $456.28$M  &  $1751.04$G  & $\bf{57.8}$ &  $456.35$M  &  $1751.34$G  &  $\bf{50.3}$  & \multicolumn{3}{c}{OOM}  \\\hline
ViT-Adapter-L    &  $\bf{570.74}$M  &  $\bf{2743.20}$G & $60.5$  &  \multicolumn{3}{c|}{NA}  &  \multicolumn{3}{c}{NA} \\
\RANKSEG &  ${584.71}$M  &  ${2747.81}$G & $\bf{60.7}$ & \multicolumn{3}{c|}{NA} & \multicolumn{3}{c}{NA}  \\
\shline
\multicolumn{10}{l}{OOM means out of memory error on $8\times32$G V100 GPUs.}
\end{tabular}
}
\label{tab:sota_seg}
\end{minipage}
\begin{minipage}[t]{1\linewidth}
\centering
\setlength{\tabcolsep}{0.1pt}
\footnotesize
\renewcommand{\arraystretch}{1.5}
\caption{\small
Combination with Mask$2$Former based on Swin-L.
}
\label{tab:sota_mask2former}
\resizebox{\linewidth}{!}
{
\begin{tabular}{ l | c | c | c | c | c | c | c | c | c | c | c | c }
\shline
\multirow{3}{*}{Method}&  \multicolumn{3}{c|}{Image semantic seg.} & \multicolumn{3}{c|}{Image panoptic seg.} & \multicolumn{3}{c|}{Video semantic seg.} &  \multicolumn{3}{c}{Video instance seg.}\\
\cline{2-13}
 &  \multicolumn{3}{c|}{ADE$20$K} & \multicolumn{3}{c|}{ADE$20$K} & \multicolumn{3}{c|}{VSPW (T=$2$)} & \multicolumn{3}{c}{YouTubeVIS $2019$ (T=$2$)} \\
\cline{2-13}
& \#params. & FLOPs  & \pbox{30mm}{mIoU($\%$)} & \#params. & FLOPs  & PQ($\%$) & \#params. & FLOPs  & mIoU($\%$) & \#params. & FLOPs  & AP($\%$) \\\shline
Mask$2$Former~\cite{cheng2021masked}   &  $\bf205.51$M &  $\bf369.02$G & $57.3$  &   $\bf205.55$M & $\bf377.98$G &  $48.1$ & $\bf205.50$M & $\bf737.38$G &  $59.4$ &  $\bf205.52$M & $\bf753.79$G & $60.4$ \\
\RANKSEG    &  $208.79$M & $369.83$G &  $\bf58.0$  & $208.83$M & $379.11$G &  $\bf48.9$ & $208.75$M & $738.14$G &  $\bf60.1$ &  $208.70$M & $754.59$G &  $\bf61.1$ \\
\shline
\multicolumn{13}{l}{T=$2$ means each video clip is composed of $2$ frames during training/evaluation and we report the GFLOPs over $2$ frames.}
\end{tabular}
}
\end{minipage}
\end{table}

\noindent\textbf{Image panoptic segmentation \& Video semantic segmentation \& Video instance segmentation}.
To verify the generalization ability of our method,
we extend RankSeg to ``rank-adaptive selected-label region classification'' and apply it to the very recent Mask$2$Former~\cite{cheng2021masked}.
According to Table~\ref{tab:sota_mask2former},
our RankSeg improves the image semantic segmentation/image panoptic segmentation/video semantic segmentation/video instance segmentation performance by +$0.7\%$/+$0.8\%$\\/+$0.7\%$/+$0.7\%$ respectively with slightly more parameters and GFLOPs.
More details about the Mask$2$Former experiments and the results of combining MaskFormer with RankSeg are provided in the supplementary material.

\section{Conclusion}
This paper introduces a general and effective rank-oriented scheme that
formulates the segmentation task into two sub-problems
including {multi-label classification} and
{rank-adaptive selected-label pixel classification}.
We first verify the potential benefits of exploiting multi-label image/video classification to improve pixel classification.
We then propose a simple joint multi-task scheme that is capable of improving various state-of-the-art segmentation methods across multiple benchmarks.
We hope our initial attempt can inspire more efforts
towards using a rank-oriented manner to solve the challenging segmentation problem
with a large number of categories.
Last, we want to point out that designing \& exploiting more accurate
multi-label image/video classification methods is a long-neglected but very important sub-problem towards more general and accurate segmentation.

\bibliographystyle{splncs04}
\bibliography{egbib}

\section{Supplementary Material}

\begin{table}[t]
\begin{minipage}[t]{1\linewidth}
\centering
\footnotesize
\setlength{\tabcolsep}{1pt}
\renewcommand{\arraystretch}{1.1}
\caption{\small Comparison with EncNet and ESSNet based on Segmenter w/ ViT-B.
}
\resizebox{1\linewidth}{!}
{
\begin{tabular}{l|c|c|c|c|c|c|c|c|c}
\shline
\multirow{2}{*}{Method}  & \multicolumn{3}{c|}{ADE$20$K}  & \multicolumn{3}{c|}{COCO-Stuff} & \multicolumn{3}{c}{COCO+LVIS}\\
\cline{2-10}
  & \#params. & FLOPs  & mIoU ($\%$) & \#params. & FLOPs  & mIoU ($\%$) & \#params. & FLOPs  & mIoU ($\%$) \\\shline
Baseline &  $102.50$M  &   $78.84$G    & $48.80$&  $102.51$M  &   $79.25$G    &  $41.85$ &  $103.37$M  &   $102.53$G  & $19.41$  \\
EncNet  &  $109.15$M  &   $84.89$G   & $49.06$  &  $109.18$M  &   $85.29$G  &  $42.81$  &  $110.90$M  &   $108.58$G  & $19.32$  \\
ESSNet  &  $\bf{101.42}$M  &   $\bf{78.05}$G  & $48.91$ &  $\bf{101.43}$M  &   $\bf{78.42}$G   &  $42.13$  &  $\bf{102.29}$M  &   $100.25$G  & $19.11$  \\
Ours     &  $109.74$M  &   $78.71$G  & $\bf{49.68}$   &  $109.70$M  &   $78.55$G  &  $\bf{44.98}$ &  $111.45$M  &   $\bf{99.59}$G  & $\bf{21.26}$  \\
\shline
\end{tabular}
}
\label{tab:comp_essnet_encnet}
\end{minipage}
\end{table}

\subsection*{A. Datasets}

\noindent\textbf{ADE$20$K/ADE$20$K-Full}.
The ADE$20$K dataset~\cite{zhou2017scene} consists of $150$ classes and diverse scenes with $1,038$ image-level labels, which is divided into $20$K/$2$K/$3$K images for training, validation, and testing.
Semantic segmentation treats all $150$ classes equally, while 
panoptic segmentation considers the $100$ thing categories and the $50$ stuff categories separately.
The ADE$20$K-Full dataset~\cite{zhou2017scene} contains $3,688$ semantic classes, among which we select $847$ classes following~\cite{cheng2021per}.

\noindent\textbf{PASCAL-Context}.
The PASCAL-Context dataset~\cite{mottaghi2014role} is a challenging
scene parsing dataset that consists of $59$ semantic
classes and $1$ background class,
which is divided into $4,998$/$5,105$ images for training and testing.

\noindent\textbf{COCO-Stuff}.
The COCO-Stuff dataset~\cite{caesar2018coco} is a
scene parsing dataset that contains $171$ semantic
classes divided into $9$K/$1$K images for training and testing.

\noindent\textbf{COCO+LVIS}.
The COCO+LVIS dataset~\cite{gupta2019lvis,jain2021scaling} is bootstrapped
from stuff annotations of COCO~\cite{lin2014microsoft} and instance annotations of LVIS~\cite{gupta2019lvis} for COCO $2017$ images.
There are $1,284$ semantic classes in total and the dataset is divided into $100$K/$20$K images for training and testing.

\noindent\textbf{VSPW}.
The VSPW~\cite{miao2021vspw} is a large-scale video semantic segmentation dataset consisting of $3,536$ videos with $251,633$ frames from $124$ semantic classes,
which is divided into $2,806$/$343$/$387$ videos with $198,244$/$24,502$/$28,887$ frames for training, validation, and testing. 
We only report the results on \texttt{val} set as we can not access the \texttt{test} set.

\noindent\textbf{YouTubeVIS}.
YouTube-VIS $2019$~\cite{yang2019video} is a large-scale video instance segmentation dataset
consisting of $2,883$ high-resolution videos labeled with $40$ semantic classes,
which is divided into $2,238$/$302$/$343$ videos for training, validation, and testing. 
We report the results on \texttt{val} set as we can not access the \texttt{test} set.

\subsection*{B. Comparison with EncNet and ESSNet}

\noindent\emph{Comparison with EncNet}.
Table~\ref{tab:comp_essnet_encnet} compares our method to EncNet~\cite{zhang2018context} based on Segmenter w/ ViT-B/$16$
and reports the results on the second and last rows.
We follow the reproduced EncNet settings in \texttt{mmsegmentation}
and tune the number of visual code-words as $64$ as it achieves the best result in our experiments.
According to the comparison results, we can see that our method significantly outperforms EncNet by +$0.62\%$/+$2.17\%$/+$1.94\%$ on ADE$20$K/COCO-Stuff/COCO+LVIS, which further verifies that exploiting rank-adaptive selected-label pixel classification is the key to our method.

\noindent\emph{Comparison with ESSNet}.
We compare our method to ESSNet~\cite{jain2021scaling} on ADE$20$K/\\COCO-Stuff/COCO+LVIS on the last two rows of Table~\ref{tab:comp_essnet_encnet}.
Different from the original setting~\cite{jain2021scaling} of ESSNet,
we set the number of the nearest neighbors associated with each pixel as 
the same value of $\kappa$ in our method to ensure fairness.
We set the dimension of the representations in the semantic space as $64$.
According to the results on COCO+LVIS, we can see that (i) our baseline achieves $19.41\%$, which performs much better than the original reported best result ($6.26\%$) in~\cite{jain2021scaling} as we train all these Segmenter models with batch-size $8$ for $320$K iterations.
(ii) ESSNet achieves $19.11\%$, which performs comparably to our baseline
and this matches the observation in the original paper that ESSNet is expected to perform better only when training the baseline method with much smaller batch sizes.
In summary, our method outperforms ESSNet by +$0.77\%$/+$2.85\%$/+$2.15\%$ across ADE$20$K/COCO-Stuff/COCO+LVIS, which further shows
the advantage of exploiting multi-label image classification over
simply applying $k$-nearest neighbor search for each pixel embedding.

{
\definecolor{person}{rgb}{0, 0.8, 0}
\definecolor{grass}{rgb}{0, 0, 0.6}
\definecolor{railing}{rgb}{0.8, 0, 0}
\begin{figure*}[htp]
\centering
\begin{subfigure}[b]{1\textwidth}
{\includegraphics[width=1\textwidth]{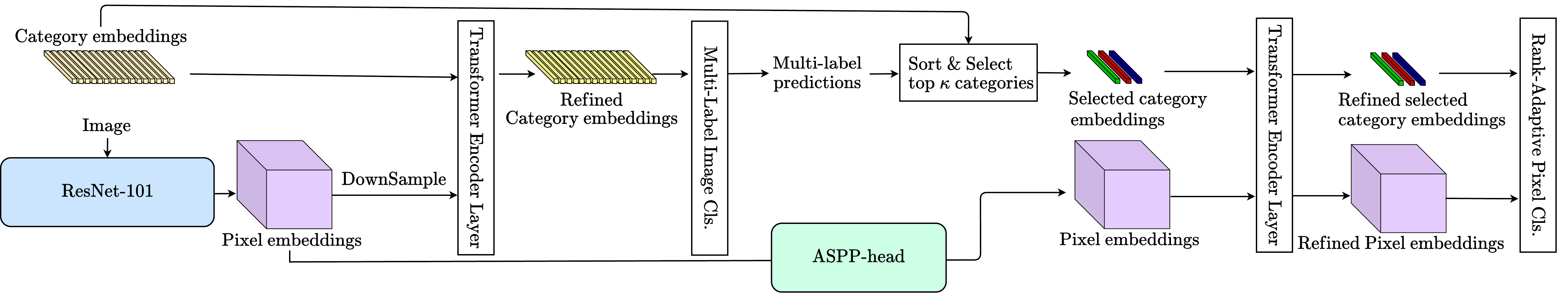}}
\caption{DeepLabv$3$ + RankSeg}
\end{subfigure}
\begin{subfigure}[b]{1\textwidth}
{\includegraphics[width=1\textwidth]{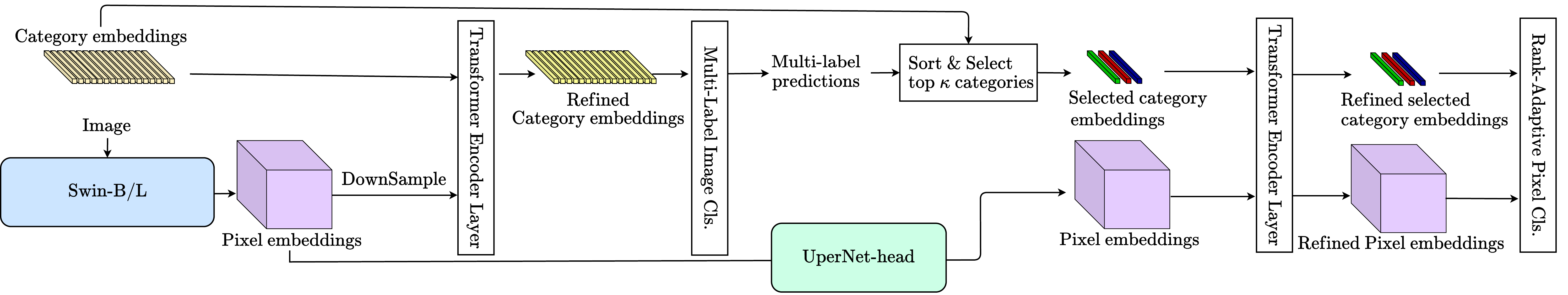}}
\caption{Swin + RankSeg}
\end{subfigure}
\begin{subfigure}[b]{1\textwidth}
{\includegraphics[width=1\textwidth]{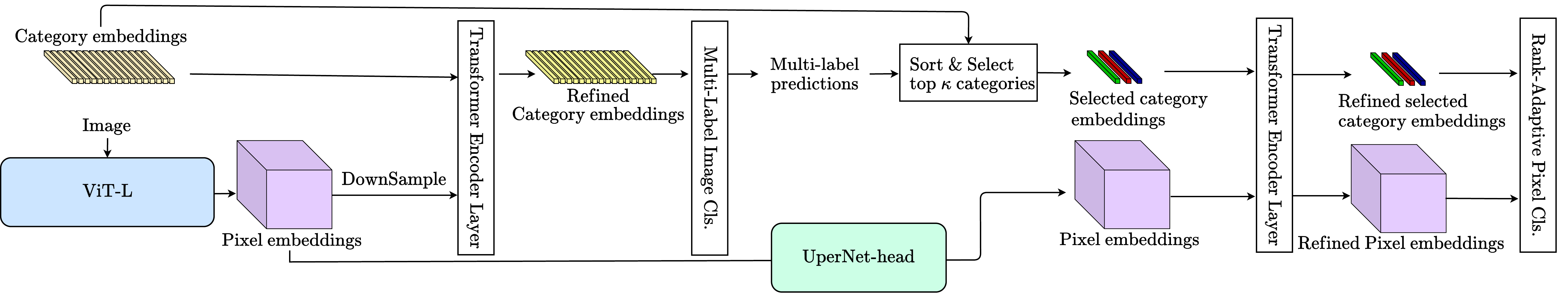}}
\caption{BEIT + RankSeg}
\end{subfigure}
\begin{subfigure}[b]{1\textwidth}
{\includegraphics[width=1\textwidth]{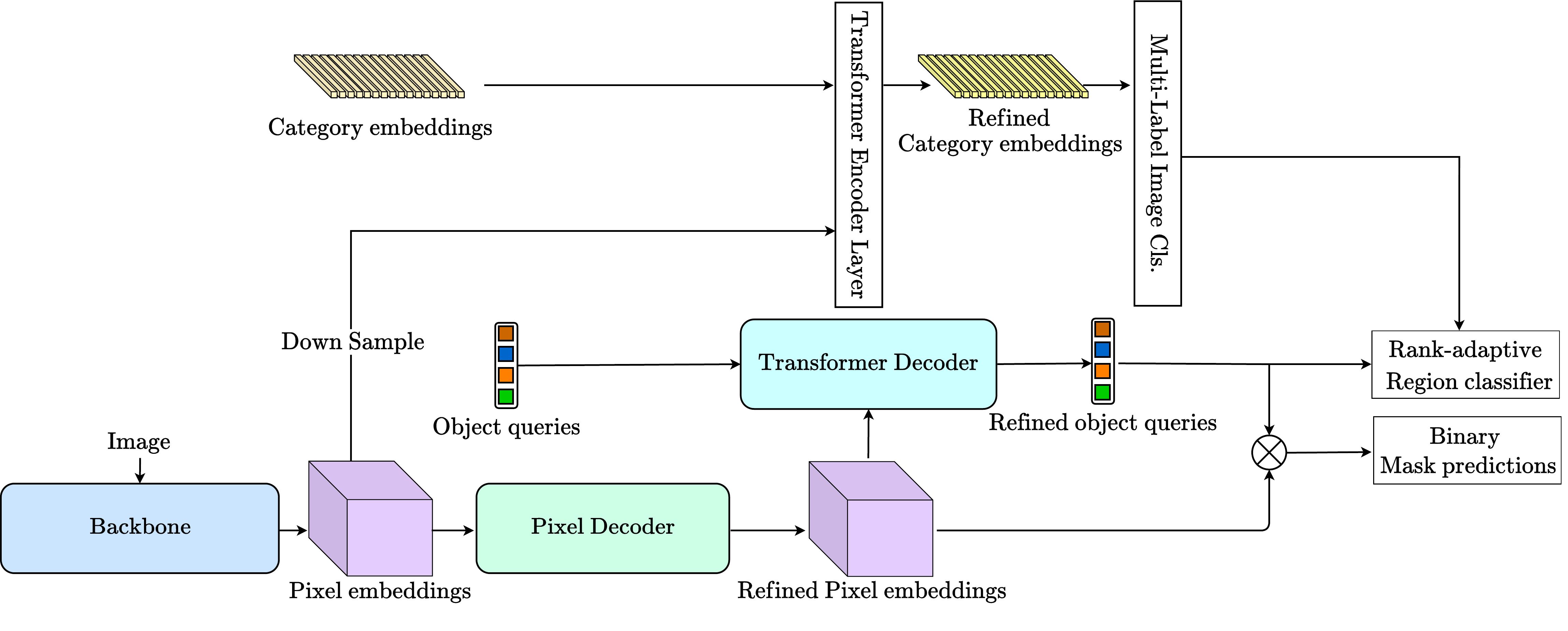}}
\caption{MaskFormer + RankSeg}
\end{subfigure}
\begin{subfigure}[b]{1\textwidth}
{\includegraphics[width=1\textwidth]{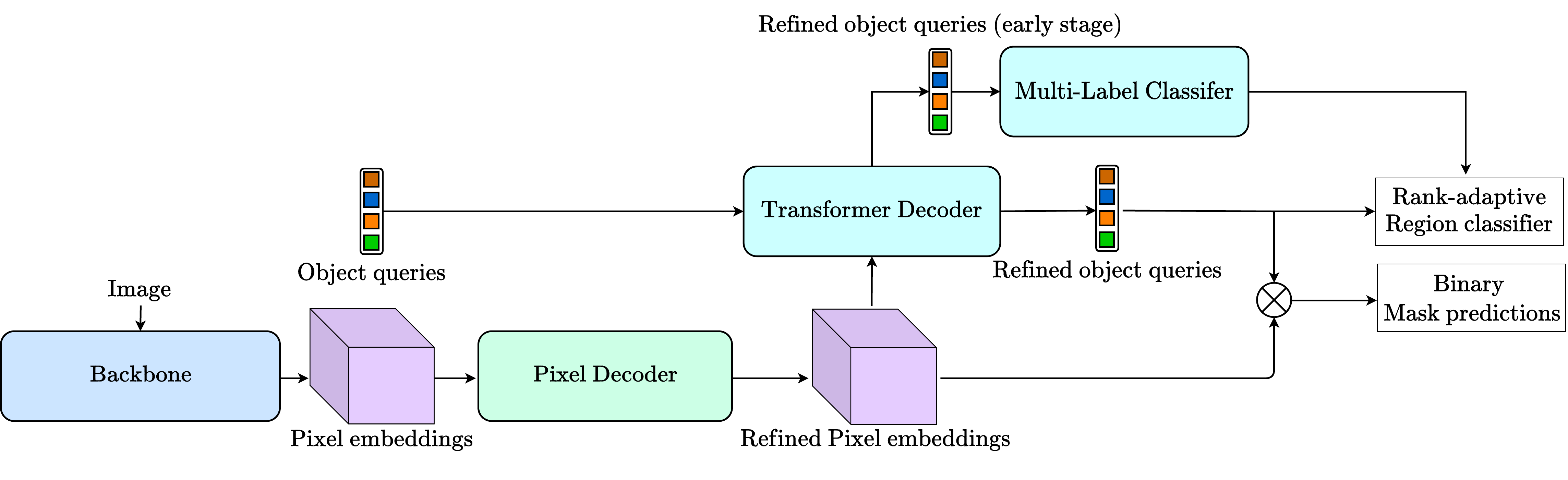}}
\caption{Mask2Former/SeMask/ViT-Adapter + RankSeg}
\end{subfigure}
\caption{\small{
The overall framework of combining our method with DeepLabv$3$~\cite{chen2017rethinking}, Swin~\cite{liu2021swin}, BEIT~\cite{bao2021beit}, MaskFormer~\cite{cheng2021per}, Mask$2$Former~\cite{cheng2021masked}, SeMask~\cite{jain2021semask}, and ViT-Adapter~\cite{chen2022vitadapter}.}}
\label{fig:dsb_framework}
\end{figure*}
}

\subsection*{C. DeepLabv$3$/Swin/\textsc{BEiT}/MaskFormer/Mask$2$Former + RankSeg}

We illustrate the details of combining our proposed joint multi-task scheme with
DeepLabv$3$/Swin/\textsc{BEiT}/MaskFormer/Mask$2$Former in Figure~\ref{fig:dsb_framework} (a)/(b)/(c)/(d)/(e) respectively.

The main difference between DeepLabv$3$/Swin/\textsc{BEiT} and the Figure~$6$ (within the main paper) is at DeepLabv$3$/Swin/\textsc{BEiT} uses a decoder architecture to
refine the pixel embeddings for more accurate semantic segmentation prediction.
Besides, we empirically find that the original category embeddings perform better than the refined category embeddings used for the multi-label prediction.
For MaskFormer and Mask$2$Former, we apply the multi-label classification scores
to select the $\kappa$ most confident categories and only apply the region classification over these selected categories, in other words,
we perform rank-adaptive selected-label region classification instead of rank-adaptive selected-label
pixel classification for MaskFormer and Mask$2$Former.
We also improve the design of Mask$2$Former + RankSeg by replacing
the down-sampled pixel embeddings with the refined object query embeddings output from the transformer decoder
and observe slightly better performance while improving efficiency.

\subsection*{D. Hyper-parameter settings on Mask$2$Former.}
Table~\ref{tab:mask2former_params} summarizes the hyper-parameter settings of experiments based on MaskFormer and Mask$2$Former.
Considering our RankSeg is not sensitive to the choice of $\kappa$/multi-label image classification loss weight/segmentation loss weight,
we simply set $\kappa$=$K$/multi-label image classification loss weight as $10.0$/segmentation loss weight as $1.0$ for all experiments and better results could be achieved by tuning these parameters.
We also adopt the same set of hyper-parameter settings
for the following experiments based on MaskFormer, SeMask, and ViT-Adapter.

\begin{table}[h]
\begin{minipage}[t]{1\linewidth}
\centering
\setlength{\tabcolsep}{0.5pt}
\footnotesize
\renewcommand{\arraystretch}{1.1}
\caption{\small
Hyper-parameter settings of Mask$2$Former + RankSeg.
}
\label{tab:mask2former_params}
\resizebox{1\linewidth}{!}
{
\begin{tabular}{ l | c | c | c | c }
\shline
\multirow{2}{*}{Method}&  \multicolumn{1}{c|}{Image semantic seg.} & \multicolumn{1}{c|}{Image panoptic seg.} & \multicolumn{1}{c|}{Video semantic seg.} &  \multicolumn{1}{c}{Video instance seg.}\\
\cline{2-5}
 &  \multicolumn{1}{c|}{ADE$20$K} & \multicolumn{1}{c|}{ADE$20$K} & \multicolumn{1}{c|}{VSPW} & \multicolumn{1}{c}{YouTubeVIS $2019$} \\
\shline
$\kappa$. & $150$ & $150$ & $124$ & $40$  \\
ml-cls. loss weight & $10$  & $10$  & $10$  & $10$ \\
seg. loss weight  & $1$  & $1$  & $1$  & $1$ \\
\shline
\end{tabular}
}
\end{minipage}
\begin{minipage}[t]{0.45\linewidth}
\centering
\setlength{\tabcolsep}{0.5pt}
\footnotesize
\renewcommand{\arraystretch}{1.3}
\caption{\small Influence of the number of the nearest neighbor within ESSNet. The class embedding dimension is fixed as $64$ by default.}
\label{tab:essnet_nn}
\resizebox{1\linewidth}{!}
{
\begin{tabular}{ l|  c | c | c | c}
\shline
\# of nearest neighbors &  $16$ & $32$ & $64$ & $100$ \\
\shline
mIoU ($\%$) & $11.07$    &   $16.19$  &  $18.45$ &  $\bf{19.11}$ \\
\shline
\end{tabular}
}
\end{minipage}
\begin{minipage}[t]{0.5\linewidth}
\centering
\setlength{\tabcolsep}{0.5pt}
\footnotesize
\renewcommand{\arraystretch}{1.1}
\caption{\small Influence of class embedding dimension within ESSNet. The number of the nearest neighbor is set as $100$ by default.}
\label{tab:essnet_dim}
\resizebox{1\linewidth}{!}
{
\begin{tabular}{ l| c | c | c | c }
\shline
Dimension & $16$ & $32$ & $ 64$ & $128$ \\
\shline
FLOPs   & $100.18$G  &  $100.21$G &  $100.25$G  &  $100.37$G \\
\shline
mIoU ($\%$)     & $19.05$  &  $19.19$ &  $19.11$  &  $\bf{19.20}$  \\
\shline
\end{tabular}}
\end{minipage}
\begin{minipage}[t]{0.37\linewidth}
\centering
\setlength{\tabcolsep}{0.1pt}
\footnotesize
\renewcommand{\arraystretch}{3}
\caption{\small Dynamic $\kappa$ with different confidence thresholds.}
\label{tab:ab_dynamic_kappa}
\resizebox{\linewidth}{!}
{
\begin{tabular}{ l| c | c | c | c }
\shline
Threshold & $0.1$ & $0.05$ & $0.02$ & $0.01$ \\
\shline
mIoU ($\%$) & $44.03$    &  $\bf{44.64}$ &  $44.54$  &  $44.56$ \\
$\bigtriangleup$   & +$2.18$ &  +$2.79$  &  +$2.69$   &   +$\bf{2.71}$ \\
\shline
\end{tabular}
}
\end{minipage}
\begin{minipage}[t]{0.62\linewidth}
\centering
\setlength{\tabcolsep}{0.5pt}
\footnotesize
\renewcommand{\arraystretch}{1.0}
\caption{\small
Combination with MaskFormer, SeMask, and ViT-Adapter.
}
\label{tab:semask}
\resizebox{\linewidth}{!}
{
\begin{tabular}{ l | c | c | c }
\shline
\multirow{2}{*}{Method}  &  \multicolumn{2}{c|}{Image semantic seg.} & \multicolumn{1}{c}{Image panoptic seg.}\\
\cline{2-4}
 &  \multicolumn{2}{c|}{ADE$20$K mIoU ($\%$)} & \multicolumn{1}{c}{ADE$20$K PQ ($\%$)} \\
\hline
Backbone    &  Swin-B  &  Swin-L  &  ResNet-$50$ \\
\shline 
MaskFormer~\cite{cheng2021per}    &  $53.9$ &  $55.6$  &   $34.7$ \\
\RANKSEG    &  $55.1$ &  $55.8$  &   $36.5$ \\\hline
SeMask~\cite{jain2021semask} & $-$ & $58.2$ & $-$ \\
\RANKSEG    &  $-$ & $58.5$ & $-$ \\\hline
ViT-Adapter~\cite{chen2022vitadapter} & $-$ & $60.5$ & $-$ \\
\RANKSEG    &  $-$ & $60.7$ & $-$ \\
\shline
\end{tabular}
}
\end{minipage}
\end{table}

{
\begin{figure}[htb]
\centering
\begin{subfigure}[b]{0.485\textwidth}
{\includegraphics[width=\textwidth]{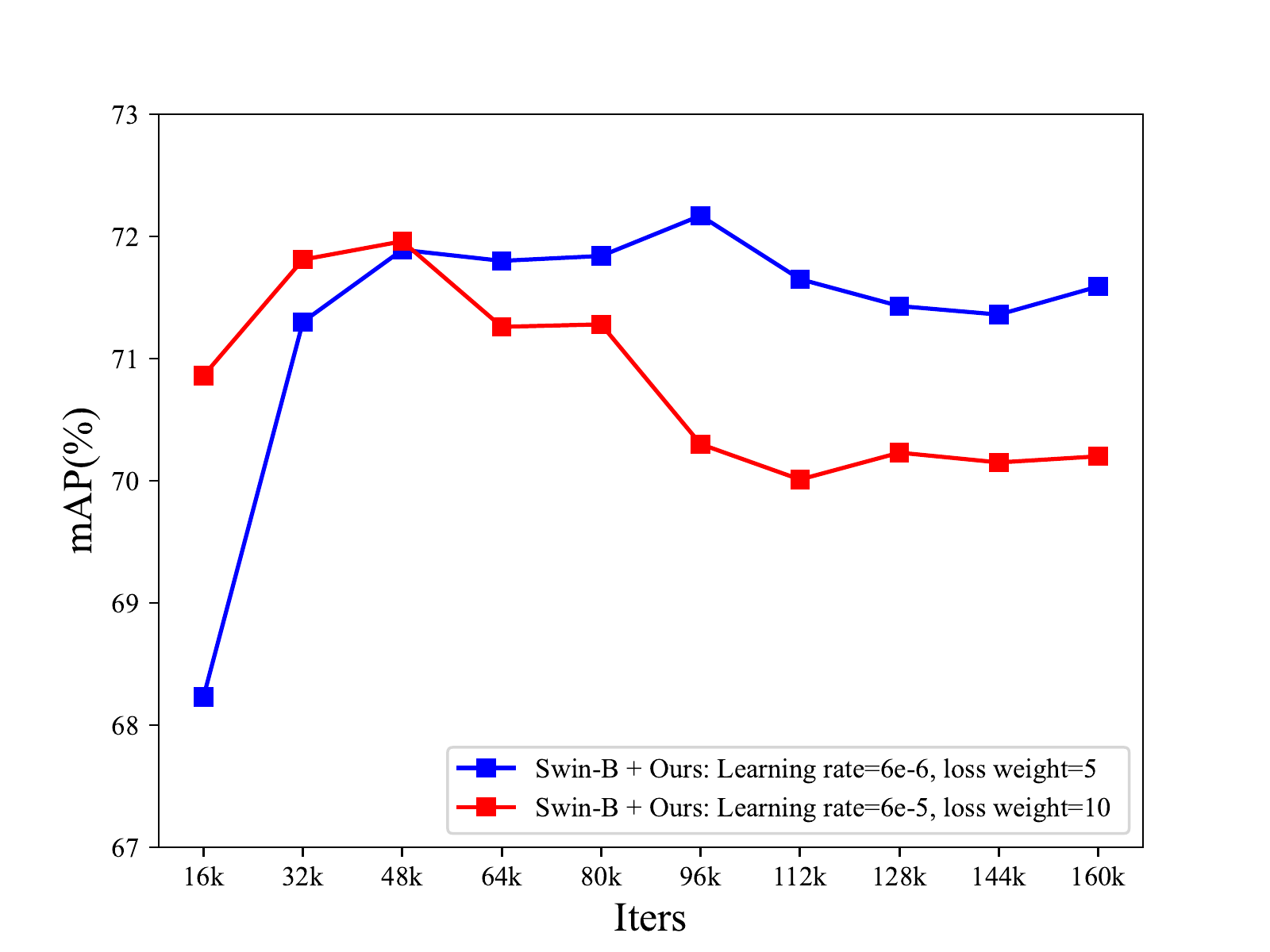}}
\caption{}
\end{subfigure}
\begin{subfigure}[b]{0.485\textwidth}
{\includegraphics[width=\textwidth]{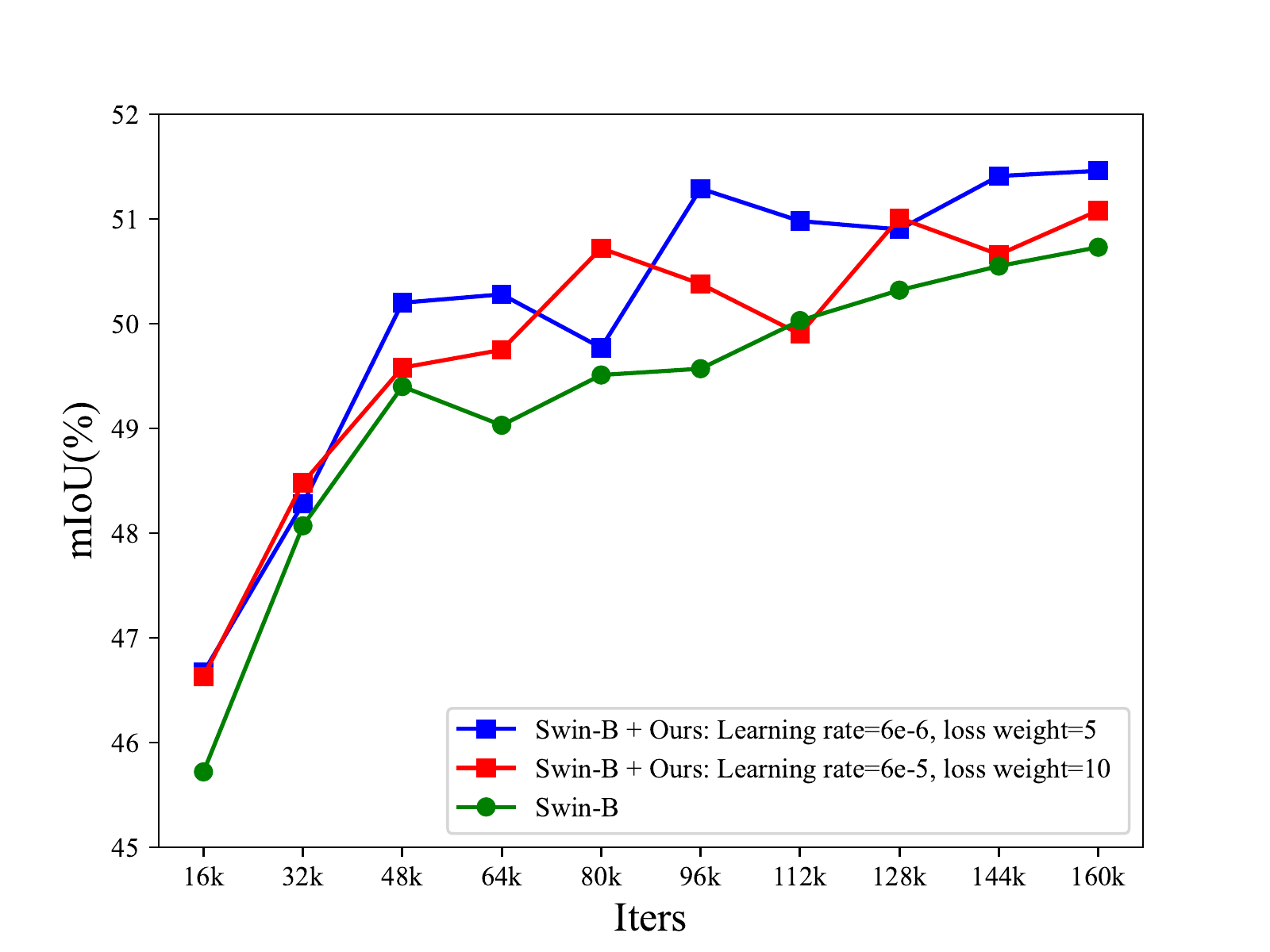}}
\caption{}
\end{subfigure}
\caption{Illustrating the curve of mAPs and mIoUs based on ``Swin'', ``Swin + RankSeg'', and ``Swin + RankSeg'' w/ smaller learning rate and loss weight on the multi-label classification head.}
\label{fig:map_curve}
\end{figure}
}

\subsection*{E. Ablation study of ESSNet on COCO+LVIS.}
We investigate the influence of the number of nearest neighbors
and the class embedding dimension in Table~\ref{tab:essnet_nn} and Table~\ref{tab:essnet_dim} based on Segmenter w/ ViT-B.

According to Table~\ref{tab:essnet_nn}, we can see that ESSNet~\cite{jain2021scaling} is very sensitive to the choice of the number of nearest neighbors. We choose $100$ nearest neighbors as it achieves the best performance.\footnote{Our method sets the number of selected categories $\kappa$ as $100$ on COCO+LVIS by default.}
Table~\ref{tab:essnet_dim} fixes the number of nearest neighbors as $100$ and compares the results with different class embedding dimensions.
We can see that setting the dimension as $32$, $64$, or $128$
achieves comparable performance.

\subsection*{F. Dynamic $\kappa$}
We compare the results with dynamic $\kappa$ scheme in Table~\ref{tab:ab_dynamic_kappa} via selecting the most confident categories,
of which the confidence scores are larger than a fixed threshold value.
Accordingly, we can see that using dynamic $\kappa$ with different thresholds consistently outperforms the baseline but fails to achieve significant gains
over the original method ($44.98\%$) with fixed $\kappa=50$ for all images.

\subsection*{G. Segmentation results based on MaskFormer and SeMask.}
Table~\ref{tab:semask} summarizes the results based on combining RankSeg with MaskFormer~\cite{cheng2021per} and SeMask~\cite{jain2021semask}.
According to the results,
we can see that our RankSeg improves MaskFormer by $1.2\%$/$1.8\%$ on ADE$20$K image semantic/panoptic segmentation tasks based on Swin-B/ResNet-$50$ respectively.
SeMask and ViT-Adapter also achieve very strong results, e.g., $58.5\%$ and $60.7\%$, on ADE$20$K with our RankSeg.

\subsection*{H. Multi-label classification over-fitting issue.}

Figure~\ref{fig:map_curve} shows the curve of multi-label classification
performance (mAP) and semantic segmentation performance (mIoU) on ADE$20$K \texttt{val} set. These evaluation results are based on the joint multi-task method ``Swin-B + RankSeg''. According to Figure~\ref{fig:map_curve} (a), we can see that the mAP of ``Swin-B + RankSeg: Learning rate=$6$e-$5$, loss weight=$10$''\footnote{The original ``Swin-B''~\cite{liu2021swin} sets the learning rate as $6$e-$5$ by default.} begins overfitting at ~$48$K training iterations and the multi-label classification performance mAP drops from $71.96$\% to $70.20$\% at the end of training, i.e., ~$160$K training iterations.

To overcome the over-fitting issue of multi-label classification,
we attempt the following strategies:
(i) larger weight decay on the multi-label classification head,
(ii) smaller learning rate on the multi-label classification head, and
(iii) smaller loss weight on the multi-label classification head.
We empirically find that the combination of the last two strategies
achieves the best result.
As shown in Figure~\ref{fig:map_curve},
we can see that using a smaller learning rate and smaller loss weight together,
i.e., ``Swin-B + RankSeg: Learning rate=$6$e-$6$, loss weight=$5$'', alleviates the overfitting problem
and consistently improves the segmentation performance.

\begin{figure}[htb]
\centering
\begin{subfigure}[b]{0.475\textwidth}
{\includegraphics[width=\textwidth,height=2.2cm]{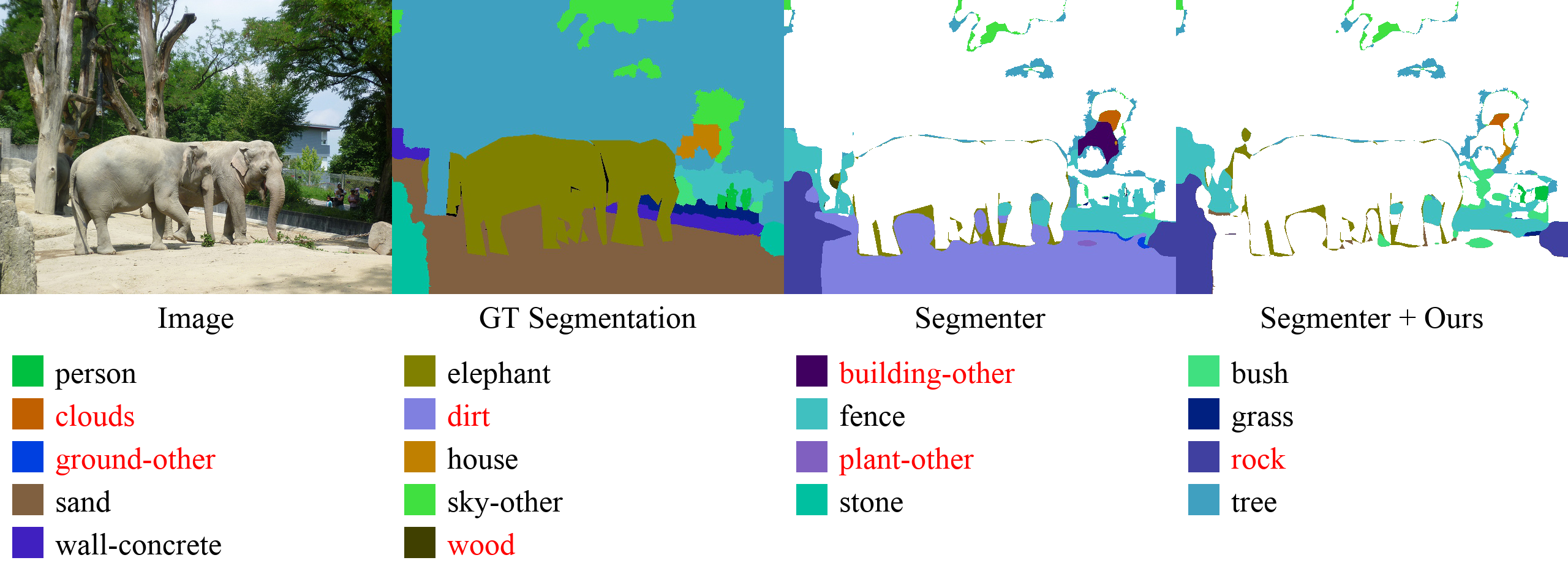}}
\caption{}
\end{subfigure}
\begin{subfigure}[b]{0.475\textwidth}
{\includegraphics[width=\textwidth,height=2.2cm]{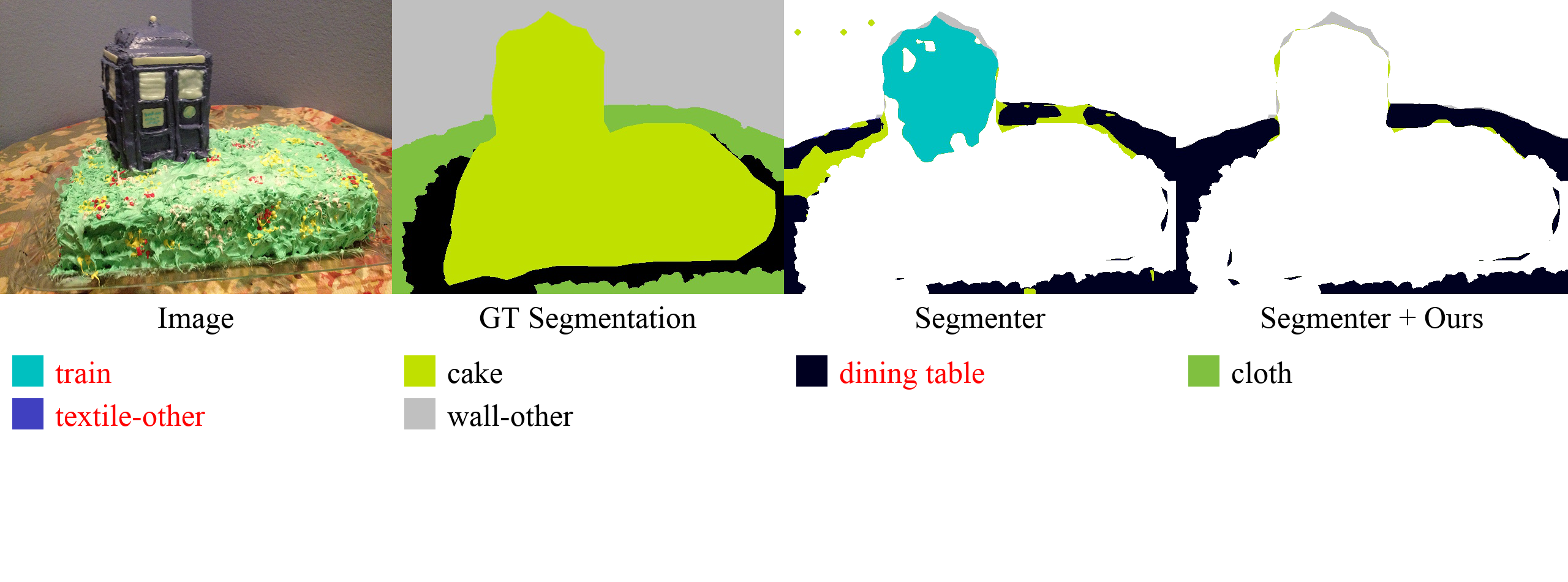}}
\caption{}
\end{subfigure}
\begin{subfigure}[b]{0.475\textwidth}
{\includegraphics[width=\textwidth,height=2.2cm]{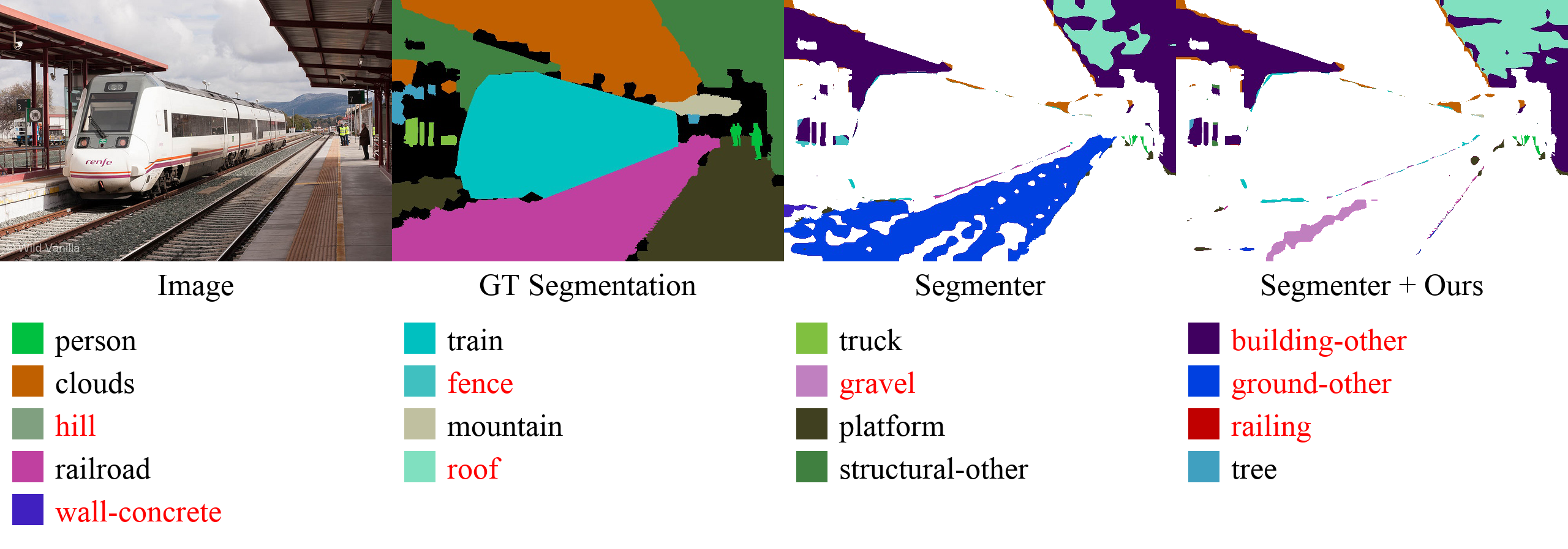}}
\caption{}
\end{subfigure}
\begin{subfigure}[b]{0.475\textwidth}
{\includegraphics[width=\textwidth,height=2.2cm]{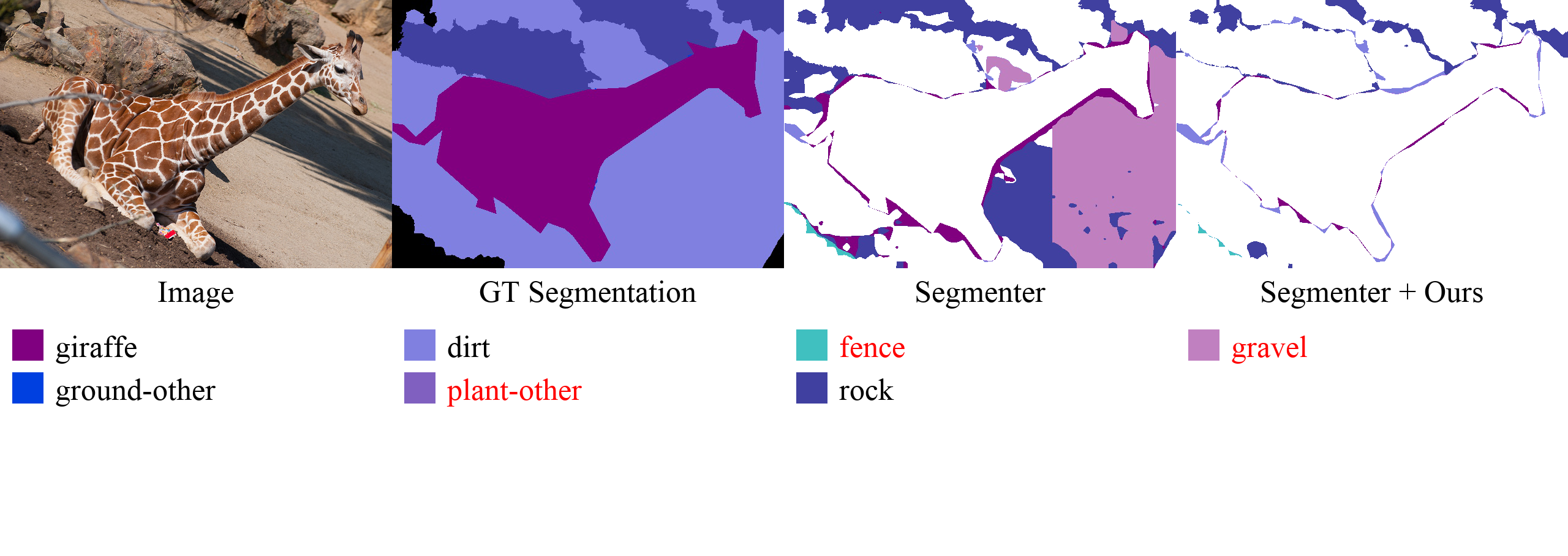}}
\caption{}
\end{subfigure}
\begin{subfigure}[b]{0.475\textwidth}
{\includegraphics[width=\textwidth,height=2.2cm]{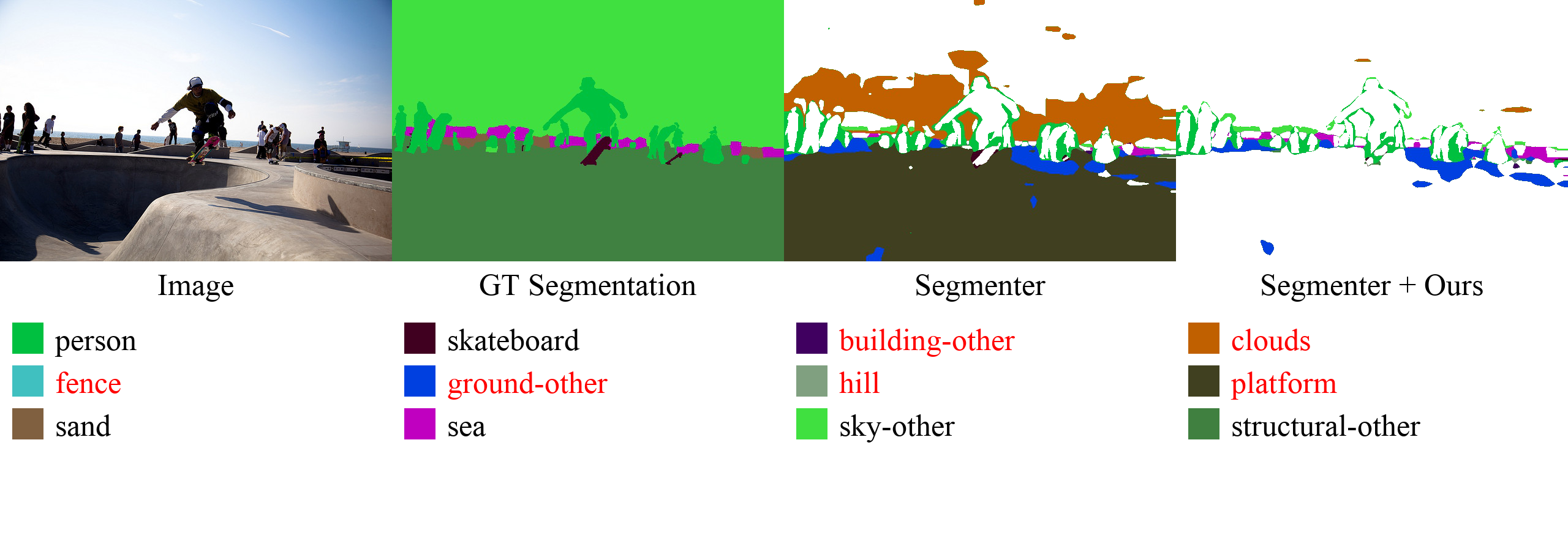}}
\caption{}
\end{subfigure}
\begin{subfigure}[b]{0.475\textwidth}
{\includegraphics[width=\textwidth,height=2.2cm]{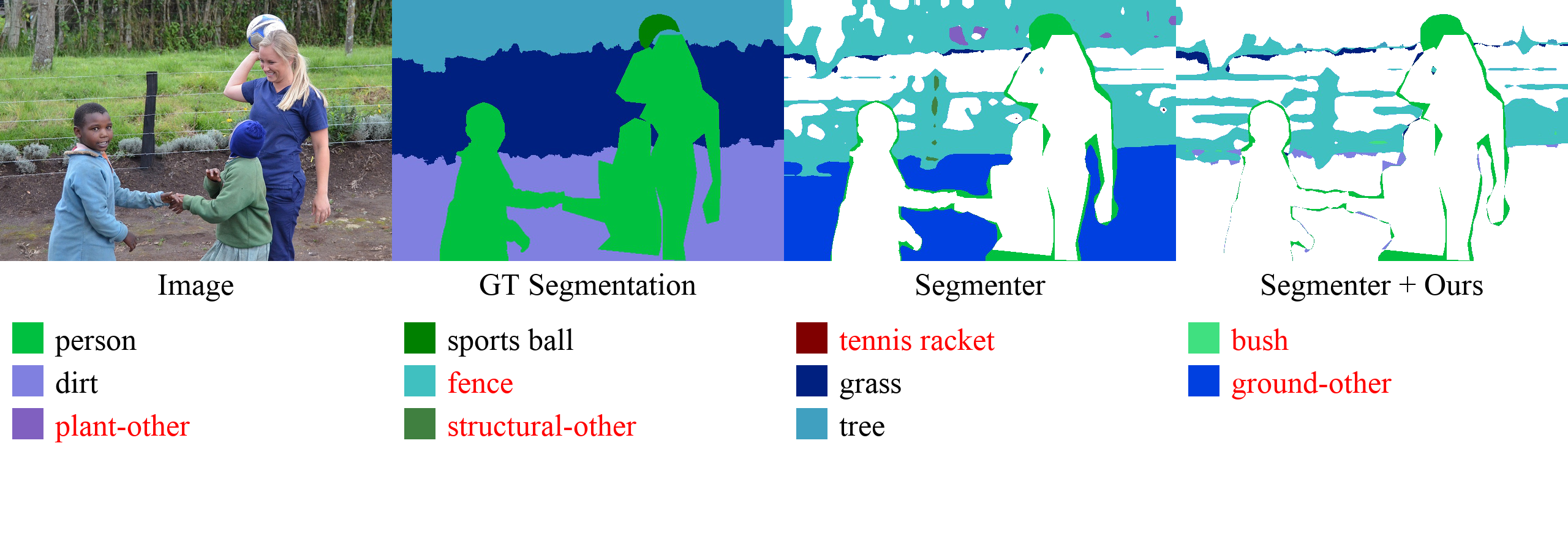}}
\caption{}
\end{subfigure}
\begin{subfigure}[b]{0.475\textwidth}
{\includegraphics[width=\textwidth,height=2.2cm]{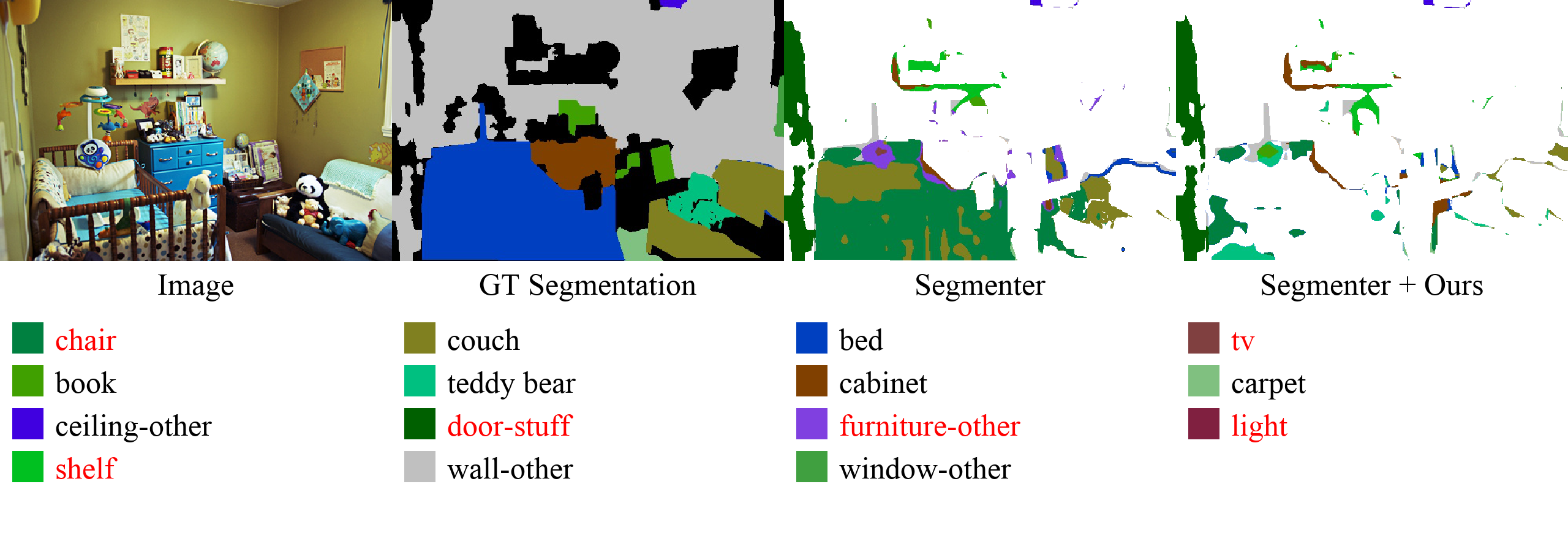}}
\caption{}
\end{subfigure}
\begin{subfigure}[b]{0.475\textwidth}
{\includegraphics[width=\textwidth,height=2.2cm]{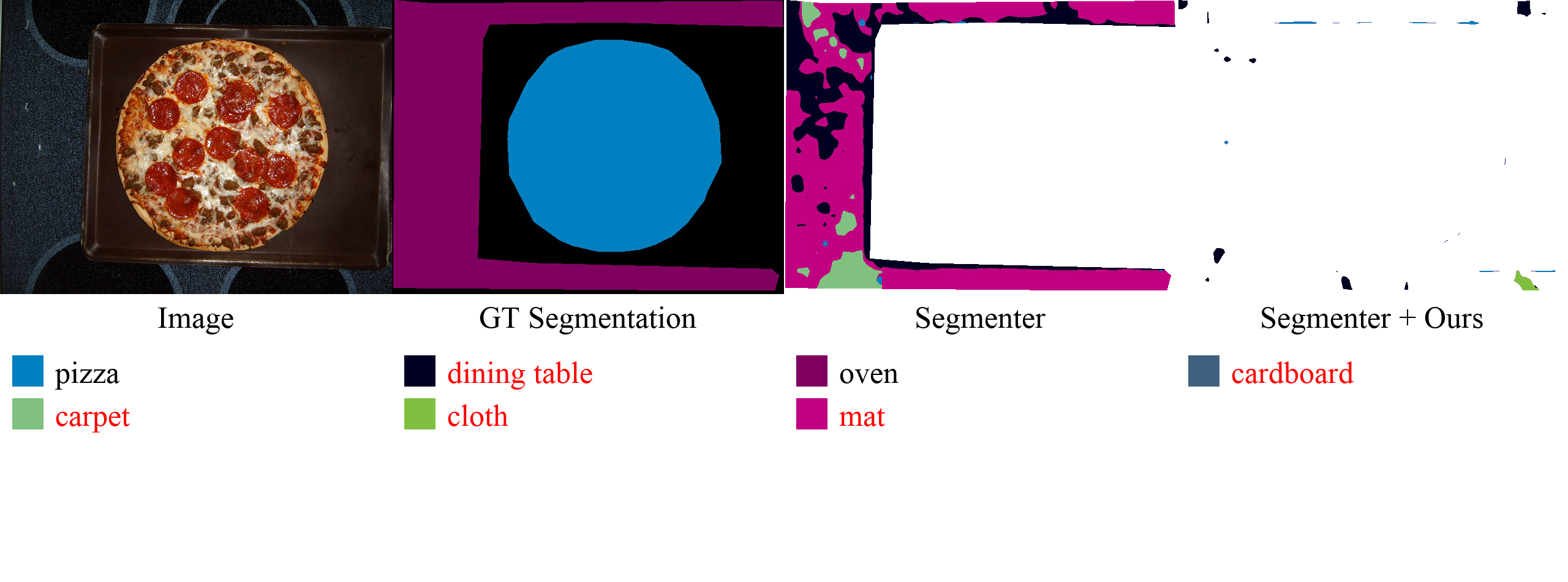}}
\caption{}
\end{subfigure}
\begin{subfigure}[b]{0.475\textwidth}
{\includegraphics[width=\textwidth,height=3cm]{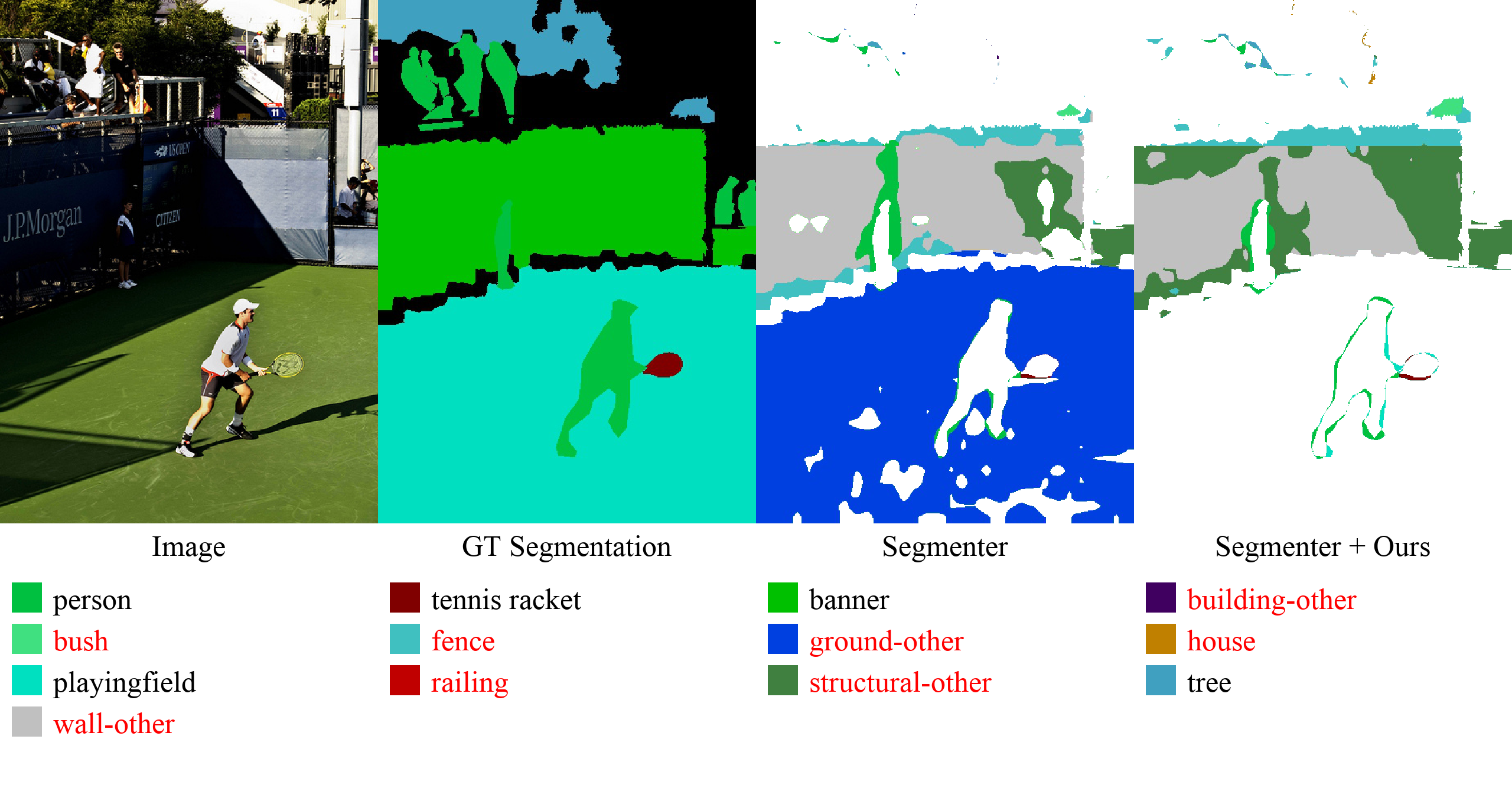}}
\caption{}
\end{subfigure}
\begin{subfigure}[b]{0.475\textwidth}
{\includegraphics[width=\textwidth,height=3cm]{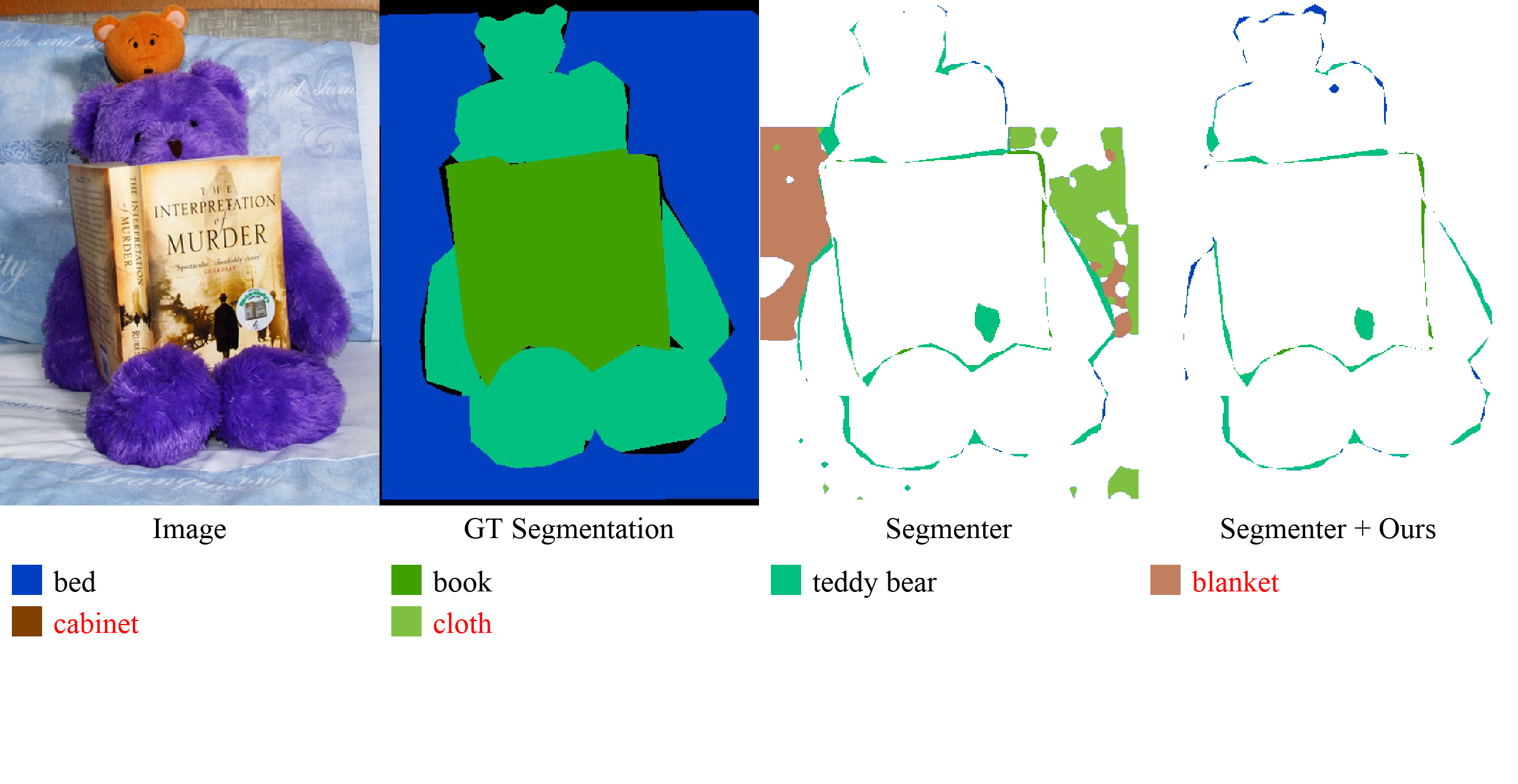}}
\caption{}
\end{subfigure}
\caption{\small{Qualitative improvements of ``Segmenter + RankSeg'' over ``Segmenter''
on COCO-Stuff \texttt{test}.
Both methods are based on the backbone ViT-B/$16$.
We mark the correctly classified pixels with white color and
the error pixels with the colors associated with the
predicted categories for the predicted segmentation results (shown on the third and fourth columns).
The names of true/{\color{red}{false}} positive categories are marked with \color{black}{black}/{\color{red}{red}} color, respectively.}
}
\label{fig:vis}
\end{figure}

\subsection*{I. Qualitative results}

We illustrate the qualitative improvement results in Figure~\ref{fig:vis}.
In summary,
our method successfully removes the false-positive category predictions of the baseline method.

\end{document}